\documentclass[conference]{IEEEtran}

\IEEEoverridecommandlockouts                              % This command is only needed if 
\usepackage{times}
\usepackage[numbers]{natbib}
\usepackage{multicol}
\usepackage[bookmarks=true]{hyperref}
\usepackage{graphicx} % for pdf, bitmapped graphics files
\usepackage{booktabs} % fancy tables
\usepackage{subfig} % self explanatory
\usepackage{algorithm} 
\usepackage[noend]{algpseudocode} % algo
\usepackage{amssymb}
\usepackage{float}
\usepackage{dblfloatfix}    % To enable figures at the bottom of page
\usepackage{flushend} % Equalize last two ref cols

\graphicspath{ {figs/} }

\hypersetup{pdfinfo={
   Title={Lightweight Unsupervised Deep Loop Closure}, %{A Lightweight Unsupervised Deep Learning Architecture for Robust Visual Place Recognition},
   Author={Nate Merrill and Guoquan Huang},
   Subject={Visual Place Recognition},
   Keywords={SLAM; Deep learning; Visual place recognition; Autoencoder; Unsupervised; Descriptor; Features; CNN; ConvNet; Feature embedding},
}}

\begin{document}

\title{ \LARGE \bf 
Lightweight Unsupervised Deep Loop Closure}

\author{Nate Merrill and Guoquan Huang %\\
%University of Delaware, Newark, DE 19716, USA
\thanks{This work was partially supported by the University of Delaware (UD) College of Engineering, UD Cybersecurity Initiative, the NSF (IIS-1566129), and the DTRA (HDTRA1-16-1-0039).} 
\thanks{The authors are with the Dept. of Computer and Information Sciences and Dept. of Mechanical Engineering, University of Delaware, Newark, DE 19716. Email: \tt{\{nmerrill,ghuang\}@udel.edu} }
}

\maketitle
%\IEEEpeerreviewmaketitle
\thispagestyle{plain}
\pagestyle{plain}

\begin{abstract}

Robust efficient loop closure detection is essential for large-scale real-time SLAM.
In this paper, we propose a novel {\em unsupervised} deep neural network architecture of a feature embedding for visual loop closure that is both reliable and compact.
Our model is built upon the autoencoder architecture, tailored specifically to the problem at hand. %loop closure in visual SLAM.
To train our network, we inflict random noise on our input data as the denoising autoencoder does, but, instead of applying random dropout, 
we warp images with randomized projective transformations to emulate natural viewpoint changes due to robot motion. % I felt 'emulate' fits better than 'capture' since we do it synthetically
Moreover, we utilize the geometric information and illumination invariance provided by histogram of oriented gradients (HOG), 
forcing the encoder to reconstruct a HOG descriptor instead of the original image.
As a result, our trained model extracts features robust to extreme variations in appearance directly from raw images, 
{\em without} the need for labeled training data or environment-specific training.
We perform extensive experiments on various challenging datasets,
showing that the proposed deep loop-closure model consistently outperforms the state-of-the-art methods
in terms of effectiveness and efficiency.
Our model is fast and reliable enough to close loops in real time with no dimensionality reduction, and capable of replacing generic off-the-shelf networks in state-of-the-art ConvNet-based loop closure systems.

\end{abstract}

%% Intro %%%%%%%%%%%%%%%%%%%%%%%%%%%%%%%%%%%%%%%%%%%%%%%%%%%%%%%%%%%%%%%%%%%%%%%%%%%%%%%%%%%%%%%%

\section{Introduction}

\begin{figure} %[t!]
        \subfloat{
          \centering
          \includegraphics[width=.235\textwidth]{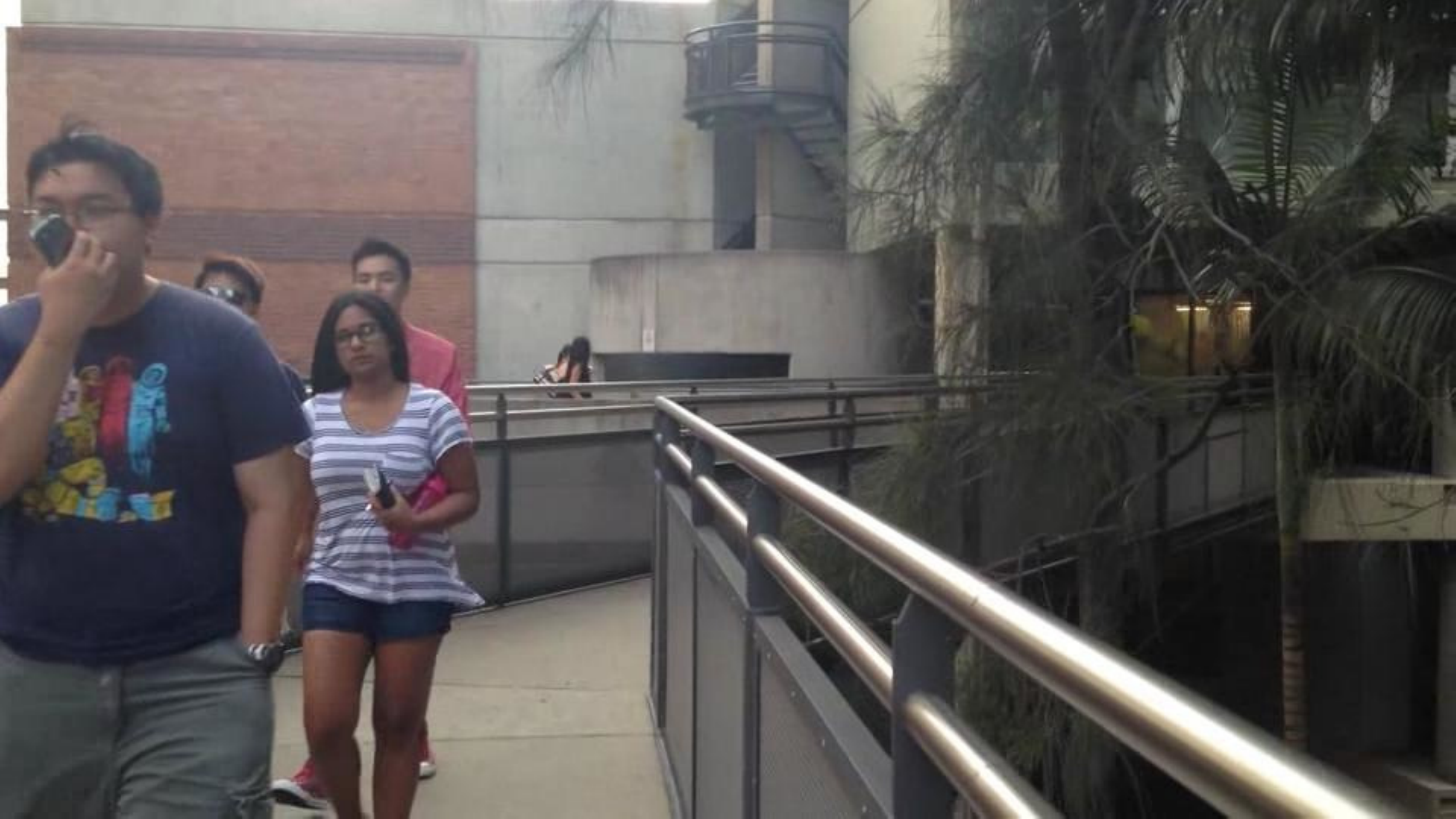}\hfill

        }
        \subfloat{
          \centering
          \includegraphics[width=.235\textwidth]{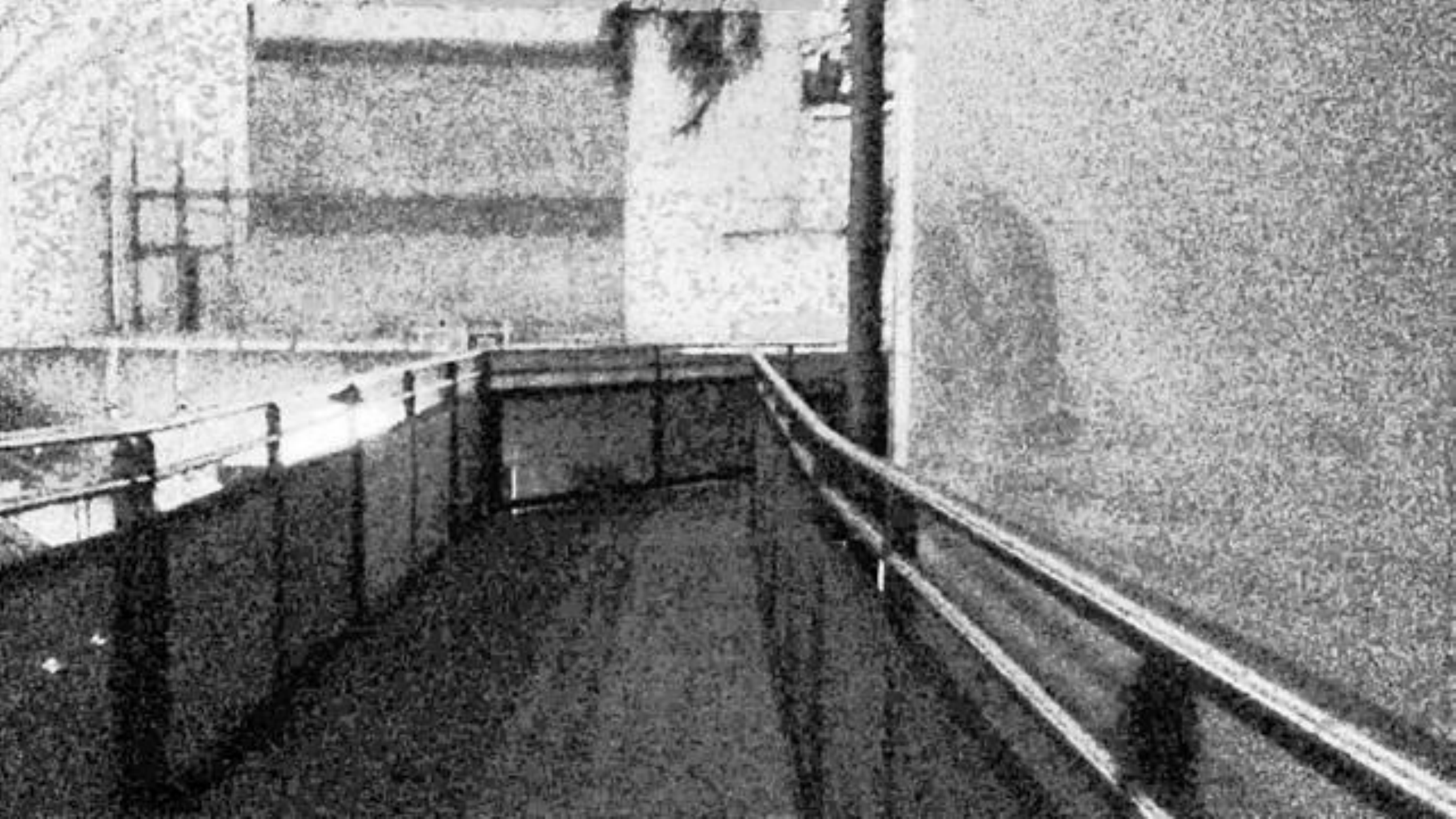}\hfill

        } \hfill\quad
        \subfloat{
          \centering
          \includegraphics[width=.235\textwidth]{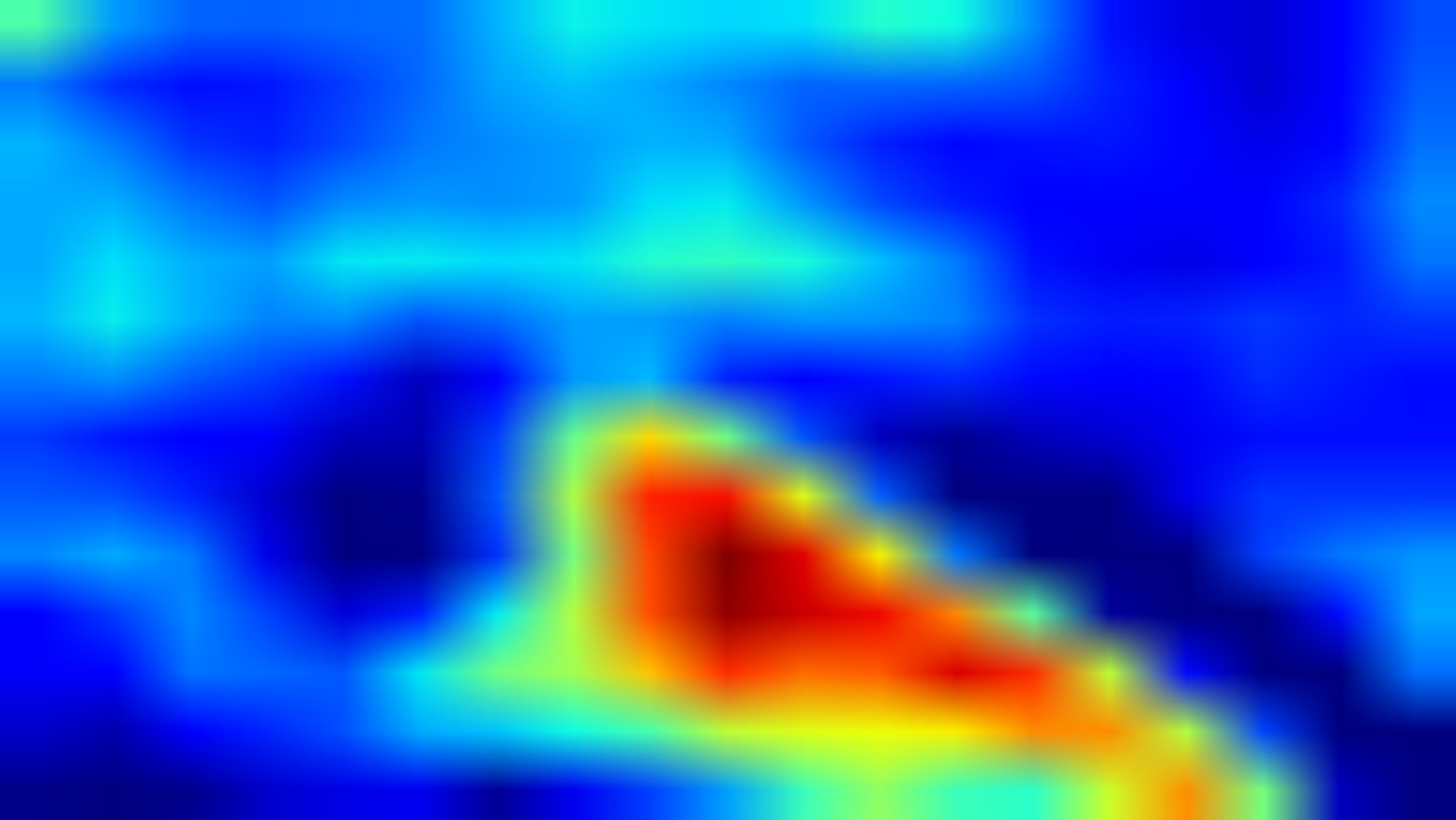}\hfill
        } 
        \subfloat{
          \centering
          \includegraphics[width=.235\textwidth]{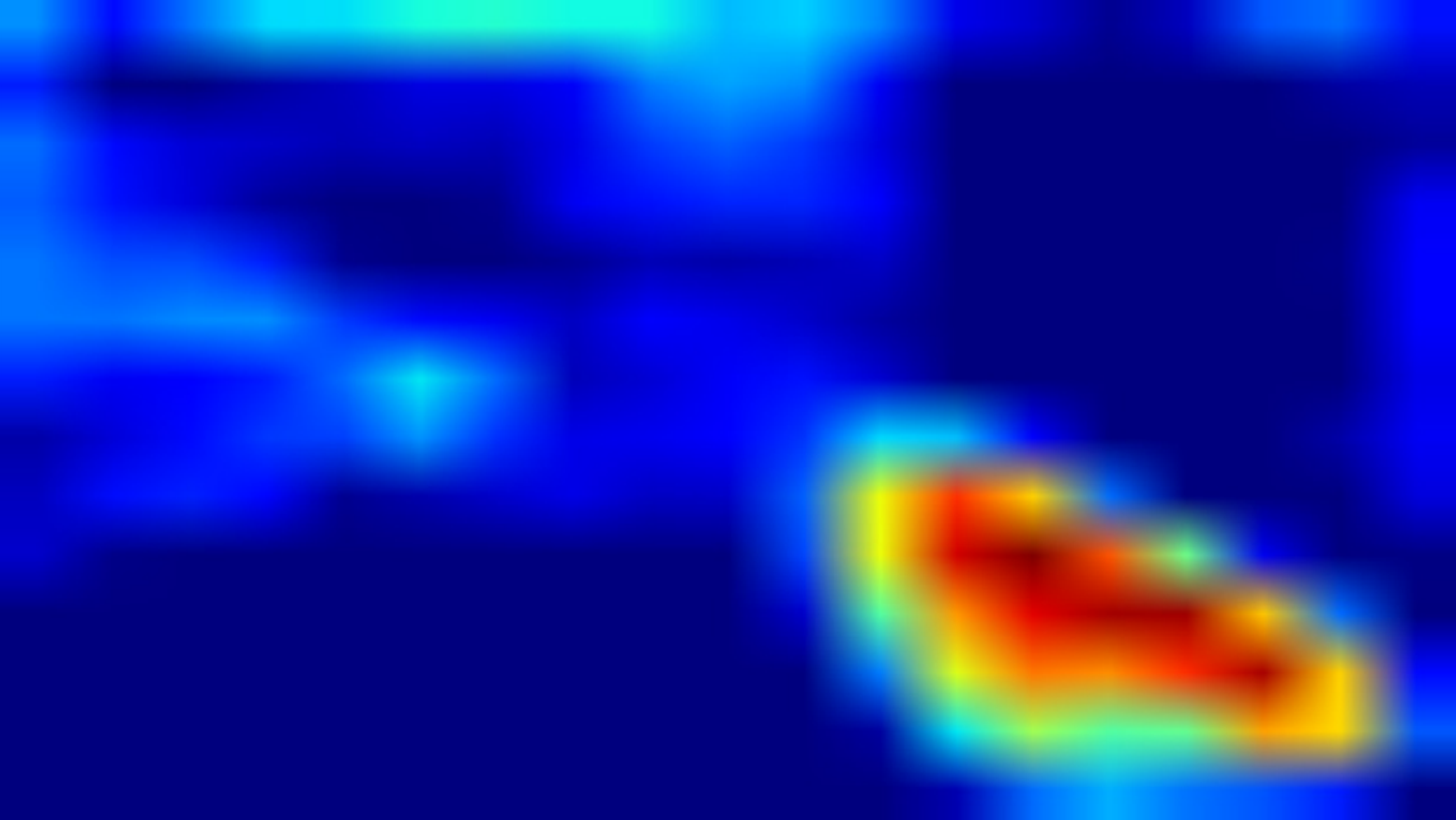}\hfill

        } \hfill
  	\caption{An example image match from the {\em Gardens Point} dataset,
which demonstrates large differences in viewpoint, dynamic objects, and illumination, as well as occlusions.
Nevertheless, with the right image as the query, our proposed method correctly retrieves the left during our experiments (see Section~\ref{sec:experiments}), while all of the tested state-of-the-art methods retrieve incorrect images.
	Below each image, the first face of the descriptor layer, before flattening, is shown.
	Evidently, these visually dissimilar images are transformed into very similar activation maps.
 }
	\label{fig:gp_example}
\end{figure}

It is critical to perform  low-latency, high-fidelity, online loop closure detection (or place recognition) for real-time visual SLAM  in order to enable bounded localization errors.
This is a challenging problem, because the visual appearance of one location at different times can change dramatically 
due to varying viewpoints, illumination, weather, and dynamic objects (see Fig.~\ref{fig:gp_example}).
Numerous algorithms have recently been developed to address these issues~\cite{Lowry2016TRO}.
% Ours uses a nearest-neighbor search as well, so I did not want it to sound like we thought this is trivial. Therefore I took out that useless point all together
Although these methods can perform well, in particular, by  incorporating  
temporal information~\cite{SeqSLAM,NaseerEtAl,GalvezTRO12,PepperellEtAl, MilfordEtAl, Bampis2017},
they may not be fast or robust enough for real-time performance in  challenging environments.

Convolutional neural networks (ConvNets)~\cite{LeCun1989} have recently become the state of the art for many vision-based classification tasks~\cite{dl-book}.
While off-the-shelf ConvNets are proven as useful feature embeddings for place recognition~\cite{Chen2014, sunderhauf_performance_2015, sunderhauf_place_2015, Razavian2014, Hou2017},
specialized networks have also been constructed and trained to further improve performance~\cite{Arandjelovic16, LOPEZANTEQUERA2017, Chen2017, gao_unsupervised_2017}. 
However, most of these ConvNet-based approaches suffer from either slow feature extraction \cite{gao_unsupervised_2017, Hou2017}, slow querying \cite{sunderhauf_performance_2015, sunderhauf_place_2015}, or the need for a large amount of labeled data for training \cite{Arandjelovic16, LOPEZANTEQUERA2017, Chen2017}.

To address the aforementioned issues, in this paper, we construct a novel autoencoder-based ConvNet for loop closure that requires very few parameters, 
and train it using public data in an {\em unsupervised} manner.
In particular, when building our autoencoder network, we exploit the advantages of classical geometric vision techniques -- 
the histogram of oriented gradients (HOG)~\cite{HOG} that offers a convenient way to compress images while preserving salient features, 
and the projective transformation (homography)~\cite{Hartley2004} that relates images with differing viewpoints.
In contrast, we also incorporate the modern stacked convolutional autoencoder into the network to be data-driven. 
%
%
%Note that 
%our model transforms raw images into their respective feature descriptors with {\em no} intermediate representation; 
%the HOG and projective transformations are used {\em only} in the training phase.
%
Consequently, the features extracted from our network are not only robust to extreme variations in appearance, 
but also lightweight and compact enough for real-time loop closing -- even for resource-constrained systems.

The main contributions of this paper are the following:
\begin{itemize}

\item We design an {\em unsupervised}, convolutional autoencoder network architecture, tailored for loop closure, 
and amenable for efficient, robust place recognition. 

\item We perform extensive comparison studies of the proposed deep loop-closure model 
against the state-of-the-art methods on different datasets.
To benefit the research community, 
we {\em open source} our code and pre-trained model used in this work along with a new dataset\footnote{The code, dataset, as well as the pre-trained model from this work are available online: \url{https://github.com/rpng/calc}.}
that captures extreme variations in viewpoint, weather, illumination, and dynamic objects in a single dataset. %, which is not often seen in public datasets.

\end{itemize}

The rest of the paper is structured as follows:
After reviewing the related work in the next section, in Section \ref{sec:methodology} we present in detail the proposed deep loop closure network,
including the network architecture, training scheme, and online usage.
%how to create the training data for our model, how to generate features, and how to create and query databases of descriptors 
% for the place recognition task
% ---highlighting any detail which might be useful for reproducibility.
The proposed approach is tested extensively in Section \ref{sec:experiments} -- both against state-of-the-art algorithms and in a real-time loop-closure setting.
Finally, we conclude the paper in Section \ref{sec:conclusions}.
%we summarize or work, and offer some insight into where it may useful in real-world applications.

%% Related Work %%%%%%%%%%%%%%%%%%%%%%%%%%%%%%%%%%%%%%%%%%%%%%%%%%%%%%%%

\section{Related Work} \label{sec:rel_work}

Due to its importance, loop closure, or place recognition, has attracted significant attention in recent years.
Many different algorithms have been introduced (see~\cite{Lowry2016TRO,Latif2017RAS} and references therein),
with variant performance characteristics in terms of complexity, robustness and efficiency.

The approaches based on bag of words (BoW), such as FAB-MAP \cite{fabmap} and DBoW2 \cite{GalvezTRO12}, 
are among the most popular for real-time visual SLAM systems (e.g., \cite{murORB2, murTRO2015, lsd_slam}).
These methods build vocabulary trees based on point features of different descriptors~\cite{Bay2008,ORB,BRIEF}, typically amenable for fast querying of matches;
%These methods extract point features of different descriptors \cite{Bay2008,ORB,BRIEF} to construct a vocabulary tree for fast query of matches. 
%which can be used for nearest-neighbor searches.
%Unfortunately, FAB-MAP suffered from run time limitations, and both of these methods 
yet, they may fail when there are large variations in appearance between images.
For this reason, 
SeqSLAM~\cite{SeqSLAM} was introduced to  utilize the information provided by image sequences to construct a better hypothesis of loop closure.
%which is a technique that can be used really to improve any place recognition pipeline.
However, this method directly compares pixel values of down-sampled images and can fail under large variations in viewpoint.
In contrast, different hand-crafted features are used in \cite{Latif2014RSS,Latif2017RAS,Zhang2016RSS}, 
where the loop closure is formulated as sparse optimization problems.

Recently, ConvNet-based approaches have risen in popularity.
%More recent works have utilized the incredible performance of convolutional neural networks to perform the place recognition task.
\citet{Chen2014} first introduced the concept of using features produced by the off-the-shelf Overfeat network~\cite{overfeat}  
as a holistic image descriptor -- shown to outperform state-of-the-art place recognition systems. 
%S\"underhauf et. al. later tested this on multiple other generic networks in 2015.
However, the descriptors extracted from such deep networks are too large to be used for real-time loop closure without approximating their similarity scores, which hinders their widespread deployment.
Since then, many similar deep learning approaches have been introduced.
For example, 
\citet{sunderhauf_place_2015} employed ConvNet features to match subregions corresponding to landmarks, improving upon the performance of the holistic image descriptors, but the authors pointed out that it was {\em nowhere} near fast enough to be used in real time.
\citet{KENSHIMOV2017124} proposed a method to omit parts of the activation maps from the neural networks in order to improve cross-seasonal place recognition.
%Aspects of older place recognition systems have also found their way into these CNN-based approaches.
%
\citet{Hou2017} combined ConvNet features with a bag of words scheme to speed up querying,
while \citet{BAI2018} combined ConvNet features with sequence searching to increase reliability.
All of these methods rely on features extracted from generic neural networks that are not trained specifically for loop closure.

Others have trained their own networks for place recognition.
For instance, 
\citet{Chen2017} compiled a large place recognition-specific dataset to train classification networks for the sole purpose of feature embedding.
NetVLAD \cite{Arandjelovic16} is an architecture that relies on geotags from Google Street View to label training images for a triplet loss scheme, 
where a triplet consists of two matching images and one non-matching image. 
\citet{LOPEZANTEQUERA2017} proposed a similar method using manually-labeled triplets, which reduces images into a single 128-dimensional vector. 
Their descriptor is shown to be useful for place recognition, 
and far more compact than that from any previous methods (e.g., \cite{Arandjelovic16}).
However, 
all of these methods rely on {\em supervised} learning -- 
requiring an immense amount of (human) effort to label images.

To address this issue, \citet{gao_unsupervised_2017} recently introduced a stacked denoising autoencoder architecture~\cite{Vincent2008} to solve the place recognition problem.
Their method is shown to perform comparably to FAB-MAP 2.0~\cite{fabmap2}, but suffers from slow feature extraction.
%Note, however, that this method completely relies on the power of the neural network to blindly learn a useful feature for place recognition.
The model employed by \citet{gao_unsupervised_2017} learns to reconstruct an image that has had random pixel values altered, but, if it is to be used for place recognition, it then has to be invariant to variations in viewpoint.
Intuitively, it would be more useful to train an unsupervised model to reconstruct an image that has been altered to mimic the viewpoint variations that it will encounter in reality.
With this observation,  in this work, 
we build upon the autoencoder concept, utilizing the multi-view geometry of homographies and the invariance of HOG, to design a novel unsupervised architecture that is both more lightweight than the previously mentioned ConvNets, and trained to compensate for the specific types of visual appearance changes that are often encountered in loop closure scenarios.

%% Methodology %%%%%%%%%%%%%%%%%%%%%%%%%%%%%%%%%%%%%%%%%%%%%%%%%%%%%%%%%%%%%

\section{Unsupervised Deep Loop Closure} \label{sec:methodology}

\begin{figure*} [!t]
	\centering
	\includegraphics[width=\textwidth]{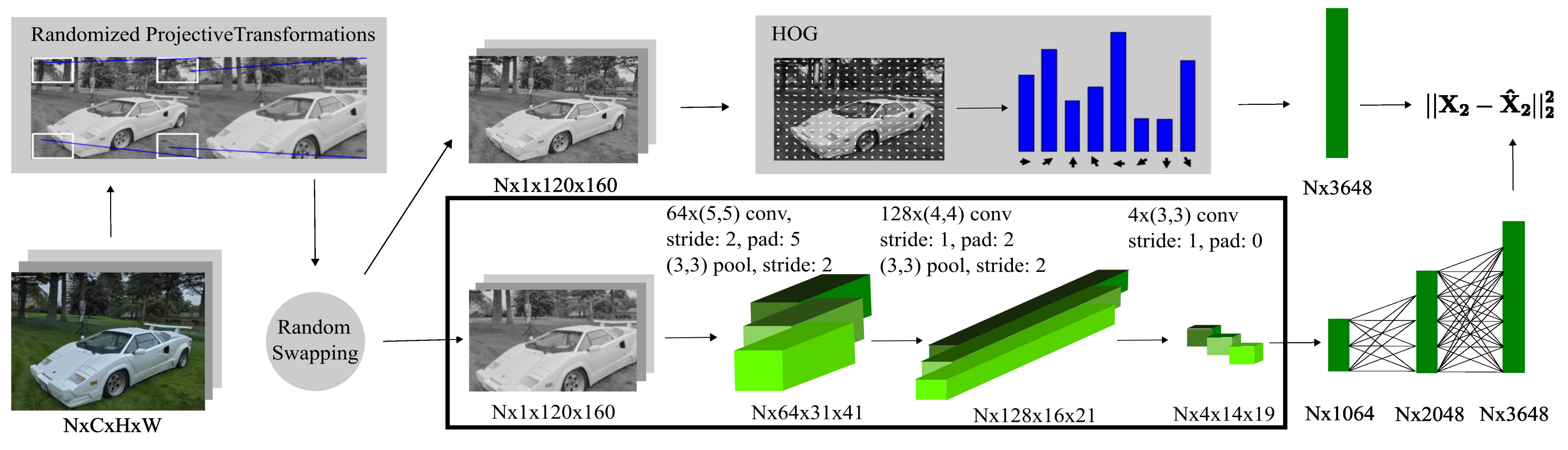}
	\caption{The training pipeline for our deep model. In this architecture, 
	%$N$ is the batch size. 
	 the projective transformations and HOG descriptors are computed only once for the entire training dataset, and the results are then written to a database to use in training. 
	 Upon deployment, the batch size $N$ is set to 1, and only the layers in the boxed area are used.
	}
	\label{fig:train_diagram}
\end{figure*}

In this section, 
we present in detail our method to construct, train, and utilize a novel autoencoder network for the loop closure task.
Our model is designed to map high-dimensional raw images into a low-dimensional descriptor space, 
which is invariant to appearance differences.
The proposed network and training scheme creates a compact robust feature embedding, while eliminating the need of image labeling.

\subsection{Design Motivation} \label{sec:design_motivation}

% To enable unsupervised learning of our deep network, a modified autoencoder architecture is employed.
%
The standard denoising autoencoder network randomly drops input values during training to mitigate the effect of noise from actual signals during testing~\cite{Vincent2008}.
Clearly, such networks do not learn the variations in images that a loop closure system will encounter, such as changes in viewpoint, illumination and so on.
Thus, the direct deployment of such autoencoders for place recognition may not be optimal.

Inspired by~\citet{sunderhauf_place_2015}, where synthetic viewpoint alterations, in the form of simple translations, were used to {\em test} their place recognition system, % Sunderhauf only used translations for testing his system. He did not train a net in this work
we employ more generalized viewpoint alterations (projective transformations) to {\em train} our deep loop-closure model.
Specifically, we inflict ``noise'' on the training inputs while modeling the natural variations due to robot motion,
thus improving the performance of the autoencoder specifically for the place recognition task.
However, these raw image pairs are not enough.
In experimentation with different network architectures,
 we constructed an autoencoder that shares the same encoding layers as the proposed model,
 but utilizes deconvolution and unpooling layers to attempt to reconstruct the other raw image from the pair.
Without any extra optimization constraints, the network learned zero vectors, which suggests that the model needs more information to map one image from the pair to the other.

HOG, by design, provides geometric information about an image.
\citet{Li2015} showed that HOG description over segmented image patches can successfully be used to match images with vastly differing appearances in a place recognition setting.
Furthermore, since HOG descriptors are fixed-length vectors for images of the same size, and can naturally be compared by the Euclidean distance, they are easily integrated into a neural network with $\ell_2$ loss.
However, since HOG relies on gradient orientation, it is not very robust to alterations in viewpoint, but, on the other hand, the image gradients are robust to illumination to some degree.
Therefore, HOG provides the prior geometric knowledge needed by our network with the added benefit of learning illumination invariance, while the random projective transformations still create the added noise required to obtain a more useful feature embedding than just HOG can provide.
Finally, it should be noted that we do not randomly place dynamic objects in the training pairs, even though our model is shown to be invariant to them (see Section \ref{sec:experiments}).
While doing so could potentially improve robustness to such occlusions, well-trained ConvNets are naturally invariant to such noise as~\citet{sunderhauf_performance_2015} observed.

\subsection{Network Architecture} \label{architecture}

Fig.~\ref{fig:train_diagram} provides a visualization of the data flow from raw images to the loss layer. 
Before training begins, every image in the set of training images $\mathcal{I}$ is converted to grayscale, resized to $120 \times 160$, and used to create an image pair (see Fig. \ref{fig:im_warp_ex} and Algorithm \ref{alg:training_data_alg}).
The HOG descriptor is computed for a randomly chosen image from each pair.
We stack all the HOG descriptors from each batch of training images, denoted by $\mathbf{X}_2$ of dimension $N\times D$, where $N$ is the batch size and $D$ is the dimension of each HOG descriptor.
The other image from the pair remains in raw form, and is stacked along with the other $N-1$ images in that training batch.
The resulting tensor denoted by $\mathbf{X}_1$ has the dimension of $N \times 120 \times 160$.

The training network aims to reconstruct $\mathbf{X}_2$ given $\mathbf{X}_1$ using only two convolution and pooling paired layers, 
one pure convolution layer, and three fully-connected layers. 
Note that every layer has an activation after it.
We use the rectified linear unit (ReLU) activation for the convolutional layers, 
while the sigmoid activation is chosen for the fully connected layers in order to better reconstruct the HOG descriptor (as it normalizes the data into $[0,1]$).
Additionally, 
since the Euclidean distance is naturally a good distance metric for HOG descriptors, 
we employ an $\ell_2$ loss function to compare $\mathbf{X}_2$ with its reconstruction $\hat{\mathbf{X}}_2$. 
Upon deployment, we drop all layers but $\mathbf{X}_1$ and the three convolution layers.
%One may notice that, as the network is fully convolutional, it is possible to change the size of the descriptor based on the input image size.
%However, in the proposed model, we extract a descriptor of dimension 1,064 from the $120 \times  160$ images, which comes from flattening the third convolutional layer.
Our model is extremely lightweight compared to the state-of-the-art models for place recognition~\cite{LOPEZANTEQUERA2017, Arandjelovic16, sunderhauf_place_2015}, taking up only 139 MB of GPU memory, 
allowing plenty of space for other processes -- even on resource-constrained low-cost platforms. % I took out 'possibly' here because we did actually test it on low-cost hardware (the Gtx 960M is available for <$100)

\subsection{Network Training} \label{sec:training_data}

\begin{figure}[t]
	\centering
	\includegraphics[width=\columnwidth]{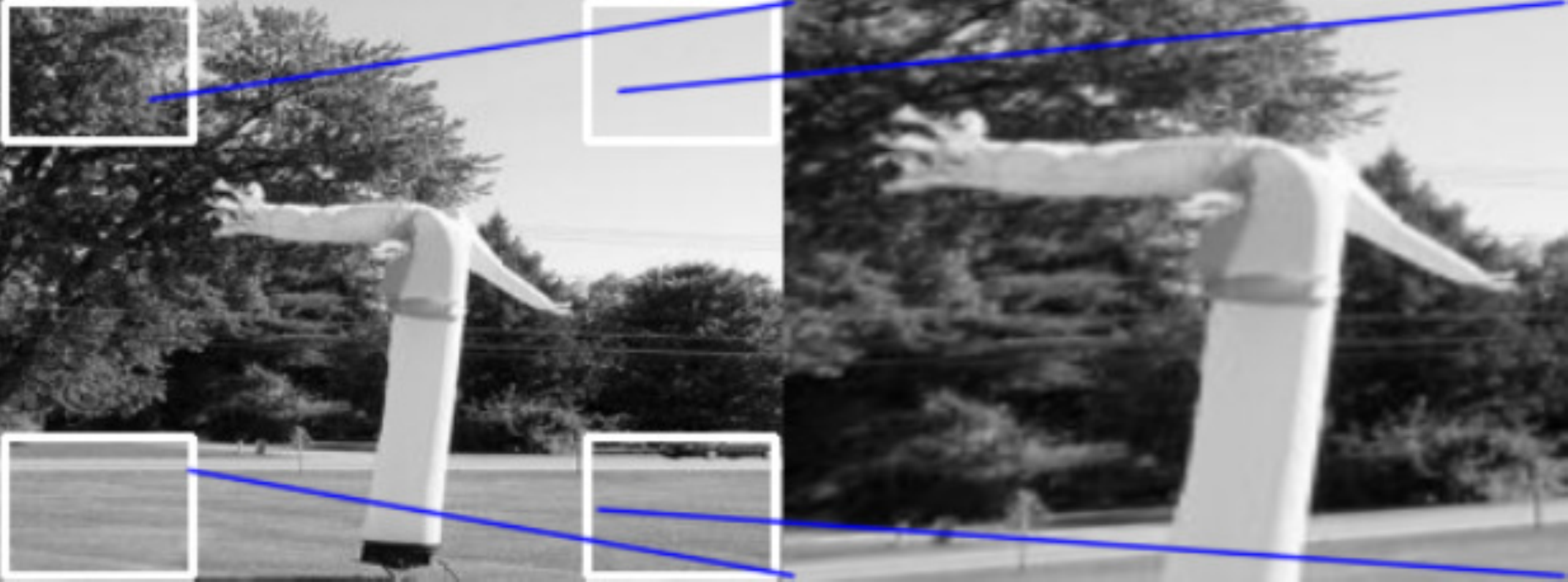}
 	\caption{An example of a possible training image pair.
		The four bounding boxes shown on the raw image (left) highlight the possible locations of each randomly selected point.
		Once the point correspondences are generated, a 2D projective transformation is calculated such that each one of those points becomes a corner of the warped image (right) after applying it.
		The randomly selected points are shown on the left, connected to their corresponding locations in the warped image shown on the right.
	 }
	\label{fig:im_warp_ex}
\end{figure}

As previously mentioned, the proposed model does not require the training images to be labeled or contain any specific information -- that is,
any image from any scene can be used in the training set to improve our model. 
To illustrate this, 
we have trained our model on the {\em Places} dataset~\cite{zhou2017places}, which has over 8 million images  originally designed for scene recognition.
Figs. \ref{fig:train_diagram} and \ref{fig:im_warp_ex} contain a few examples of images from this dataset.
While the majority of the images in the {\em Places} dataset are unrelated to any scene that a loop closure algorithm may encounter, the sheer number of images leads to improved performance over training on smaller datasets.
%Here we describe the methodology to transform a set of random images into data that is useful for training our model.
Algorithm~\ref{alg:training_data_alg} outlines the main steps of utilizing such a dataset to create $\mathbf{T}_1$ and $\mathbf{T}_2$, the large tensors from which $\mathbf{X}_1$ and $\mathbf{X}_2$ are sampled during every iteration of stochastic gradient descent. 
%The batches are sampled in-order since the data is initially randomly shuffled.

\begin{algorithm} [!t]
	\caption{Generating Training Data}\label{alg:training_data_alg}
	\textbf{input:} $\mathcal{I}$: A set of grayscale training images, resized to $\textit{H
} \times \textit{W}$

	\textbf{output:} $ \mathbf{T}_1 \in \mathbb{R}^{\textit{M} \times \textit{H} \times \textit{W}} $ and $ \mathbf{T}_2 \in \mathbb{R}^{\textit{M}\times\textit{D}}$ 

	\textbf{define:} $ \textit{rand}(\mathcal{A}) $ as a map from set $\mathcal{A}$ to one of its elements, chosen at random

	\begin{algorithmic}[1]
		\State $ \textit{W, H, D, M} \gets 160, 120, 3648, |\mathcal{I}| $ % I had to remove the comma from 3648 since it looks like a 3 and a 648 here
		\State $ \mathbf{T}_1 \gets \mathbf{0}_{\textit{M} \times \textit{H} \times \textit{W}} $ %\Comment{Raw grayscale images}
		\State $ \mathbf{T}_2 \gets \mathbf{0}_{\textit{M} \times \textit{D}} $% \Comment{HOG Descriptors}
		\State $ \mathbf{P}_c \gets ((0,0),(0,\textit{H}),(\textit{W},0),(\textit{W,H})) $ %\Comment{Image corners}
		\For{$ \textit{i} \in \mathbb{N} \cap [1,\textit{M}] $}
			\State $ \mathbf{I} \gets \textit{rand}(\mathcal{I}) $
			\State $ \mathbf{P}_r \gets \textit{randFourPts}(\textit{W,H}) $% \Comment{Bounds shown in Fig. \ref{fig:im_warp_ex}}
			\State $ \mathbf{H}_p \gets \textit{estimateHomography}(\mathbf{P}_r, \mathbf{P}_c)  $
			\State $ \mathbf{I}_w \gets \textit{transform}(\mathbf{I}, \mathbf{H}_p) $
			\If{$\textit{rand}(\{0,1\})$} %\Comment{Randomly switch $\mathbf{I}$ and $\mathbf{I}_w$}
				\State $ \textit{swap}(\mathbf{I}, \mathbf{I}_w) $
			\EndIf
			\State $ \mathbf{T}_1^{(\textit{i})} \gets \mathbf{I} $ 
			\State $ \mathbf{T}_2^{(\textit{i})} \gets \textit{calcHOG}(\mathbf{I}_w) $
		\EndFor
	\end{algorithmic}
\end{algorithm}

Given an image $\mathbf{I} \in \mathcal{I}$, we would like to automatically generate $\mathbf{I}_w$, which is of the same scene as $\mathbf{I}$ from a different viewpoint; this effect is achieved by applying a randomized 2D projective transformation matrix, $\mathbf{H}_p \in \mathbb{R}^{3 \times 3}$, to every pixel location in $\mathbf{I}$.
To obtain this matrix, four points are randomly selected within the bounding boxes along the corners in the image $\mathbf I$ (see Fig.~\ref{fig:im_warp_ex}); $\mathbf{H}_p$ is then calculated to warp $\mathbf{I}$ such that those four points become the four corners of $\mathbf{I}_w$.
We choose each bounding box of the point selection to be $H/4 \times W/4$ in order to avoid excessive distortion of $\mathbf{I}_w$, while still warping it enough to emulate a new perspective of the scene.
Since every $\mathbf{I}_w$ appears zoomed in compared to $\mathbf{I}$, we randomly choose which of the images out of every pair to place into $\mathbf{T}_1$, avoiding unnecessary training biases.

We employ a HOG descriptor with large strides and a small window size to reduce the dimension of one of the images in each training pair --
%
%Specifically, we use  $16 \times 32$ window size, $16 \times 16$ block size, $16 \times 16$ block stride, $8 \times 8$ cell size, and 2 bins to compute the reduced-size HOG, 
mapping an image $\mathbf I \in \mathbb{R}^{120 \times 160}$ of 19,200 pixels to $\mathbb{R}^{3,648}$.
While this particular HOG descriptor may not be very informative for place recognition because of its aggressive data compression, 
it helps the autoencoder model to learn a good image encoding as mentioned in Section~\ref{sec:design_motivation}.
To construct and train our model, we utilize the {Caffe Deep Learning Library}~\cite{jia2014caffe} due to its efficiency.
%
%$N$ is set to 768 during training, but we use two GPUs to create an effective batch size of 1,536.
%From cross validation, we choose a fixed learning rate of 0.0009, and an iteration count of 220,000.
We train our model for roughly 42 epochs with a fixed learning rate of $9\times10^{-4}$.
Based on~\citet{AlexNet}, we choose a momentum of 0.9, and weight decay of $5\times10^{-4}$. % Prior work makes it sound like we authored that paper
%A stochastic gradient descent solver is used to optimize our model.
%

\subsection{Online Use} \label{sec:online_use}

Once our model is trained, upon its deployment for online use, 
we create a database of the descriptors extracted by our model and later query it to find loop closure candidates.
While K-D trees \cite{Bentley1975} are a popular means to create such databases for nearest-neighbors searches, there is no speed up over a linear search for 1,064-dimensional vectors -- even when the search is approximated \cite{flann_pami_2014}. 
For this reason, we use the simple linear search method.
Additionally, as the descriptors are compact enough, their similarity can be calculated directly with no dimensionality reduction.

%
%Specifically, 
%in the nearest-neighbor search with a query image $\mathbf{I}_q$, we first calculate the normalized descriptor vector $\hat{\mathbf{d}}_q$,
%and then query the database, finding $i$, and thus, $\hat{\mathbf{d}}_i$, which maximizes the cosine similarity score,
%$\mathcal{S}_{q,i} = \hat{\mathbf{d}}_{q}^{\intercal} \hat{\mathbf{d}}_i$.
%
%
We seek to emphasize that our method of creating and querying the database with the descriptors extracted from our model is simple but effective;
albeit, we are able to achieve faster-than-real-time querying speed with minimal memory usage (see Section~\ref{sec:runtime-eval}).
Furthermore, since many new ConvNet-based place recognition systems~\cite{sunderhauf_place_2015, KENSHIMOV2017124, Hou2017, BAI2018} rely on features from bulky off-the-shelf networks, our light-weight model can potentially be utilized in many of these systems to achieve speedups with competative accuracy (see Section \ref{sec:improving_performance}).

%% Experiments %%%%%%%%%%%%%%%%%%%%%%%%%%%%%%%%%%%%%%%%%%%%%%%%%%%%%%%%%%%%%%%

\section{Experimental Results} \label{sec:experiments}

To validate the proposed unsupervised deep loop closure model,
we have performed extensive comparison studies on various datasets with the state-of-the-art approaches as well as other benchmarks where applicable.
While runtime is used as the criterion for evaluating efficiency,
we utilize the precision-recall curve, a standard method to evaluate binary classification, to quantify effectiveness.
%
%Specifically, precision is defined as $\frac{t_p}{t_p + f_p}$, where $t_p$ and $f_p$ are the numbers of true and false positives, respectively, 
%and the recall is given by $\frac{t_p}{t_p + f_n}$, where $f_n$ is the number of false negatives.
%These values are generated by varying the threshold of the similarity score to declare a positive classification, 
%which is a loop closure hypothesis in these experiments. % I took this out, because this is common knowledge
%
While there are many ways to interpret a precision-recall curve, we  primarily use: %the following two ways:
(i) the area under curve (AUC), where a higher AUC is desired;
and (ii) the maximum recall rate with 100\% precision,  denoted by $r$, where again a higher value is desired.
This can be observed visually in any precision-recall curve, as it will be the recall rate where the precision first dips down from 1.0.
%This value will be referred to as $r$ from here on out for brevity.
By observing both of these values, we obtain a comprehensive picture about how well the considered algorithms can generalize; however, the $r$ value is slightly more desirable in practice, since one binary classifier can have non-perfect precision for all recall rates despite a high AUC.

For the results presented below,  we compare the proposed approach with the following:
(i) {\bf Autoencoder}: A traditional denoising convolutional autoencoder.
 This model has the same encoding layers as our proposed model upon deployment, but, instead of reconstructing HOG descriptors of warped images, it utilizes deconvolution and unpooling layers to reconstruct the original image and is subject to random dropout during training.
(ii) {\bf LA}: The model from \citet{LOPEZANTEQUERA2017}, which has comparable efficiency as our {\em unsupervised} model while requiring labeled data for training (i.e. {\em supervised}), making any retraining difficult.
(iii) {\bf DBoW2}: We use the DBoW2 vocabulary tree from the state-of-the-art ORB-SLAM~\cite{murTRO2015, murORB2}. 
(iv) {\bf AlexNet}: \citet{sunderhauf_performance_2015} found AlexNet \texttt{conv3} to be the most robust layer for place recognition; however, it was also noted that the 64,896-dimensional vector produced was too large to perform real-time database queries.
Therefore, we apply Gaussian random projection (GRP)~\cite{Dasgupta2000, Bingham2001} as in~\cite{sunderhauf_place_2015} to compress the \texttt{conv3} layer to the same size as the descriptors from the proposed model.
In our tests, we use the AlexNet trained by BVLC.
(v) {\bf HOG}: Although the 3,648-dimensional HOG descriptor is used to train our model,
we include this comparison merely to show that our model is able to learn a better feature than the original reduced HOG,
rather than to show the ability of HOG as descriptors for place recognition.
Note that for all of these methods, we use a single nearest-neighbor linear search in order to purely compare the ability of each descriptor to match places.
At last,
it should be pointed out that 
in all the following experiments, 
our approach uses the {\em same} model  trained on a completely {\em different} dataset from the testing datasets, 
showing that the proposed deep loop closure network does not require environment-specific training.

\begin{figure} [!t]
	\centering
	\subfloat{
	  \centering
	  \includegraphics[width=.475\columnwidth]{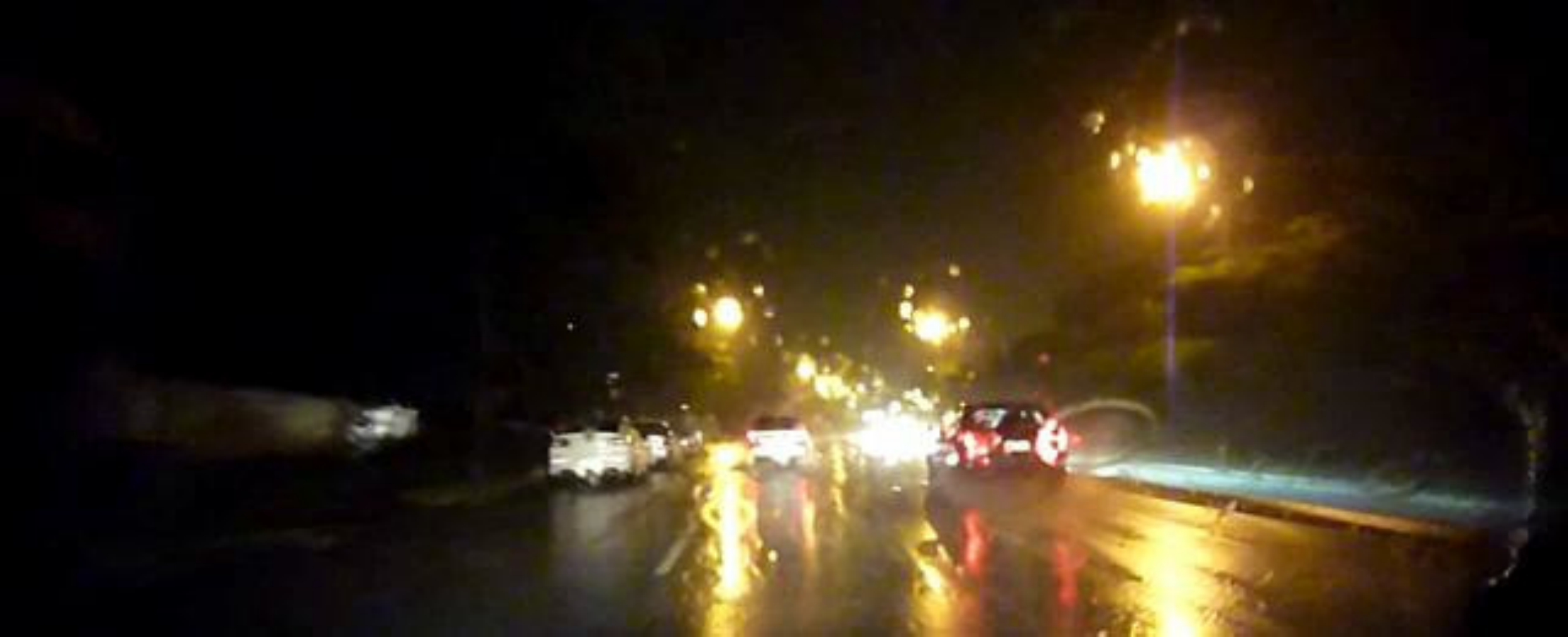}
	} %\quad
	\subfloat{
	  \centering
	  \includegraphics[width=.475\columnwidth]{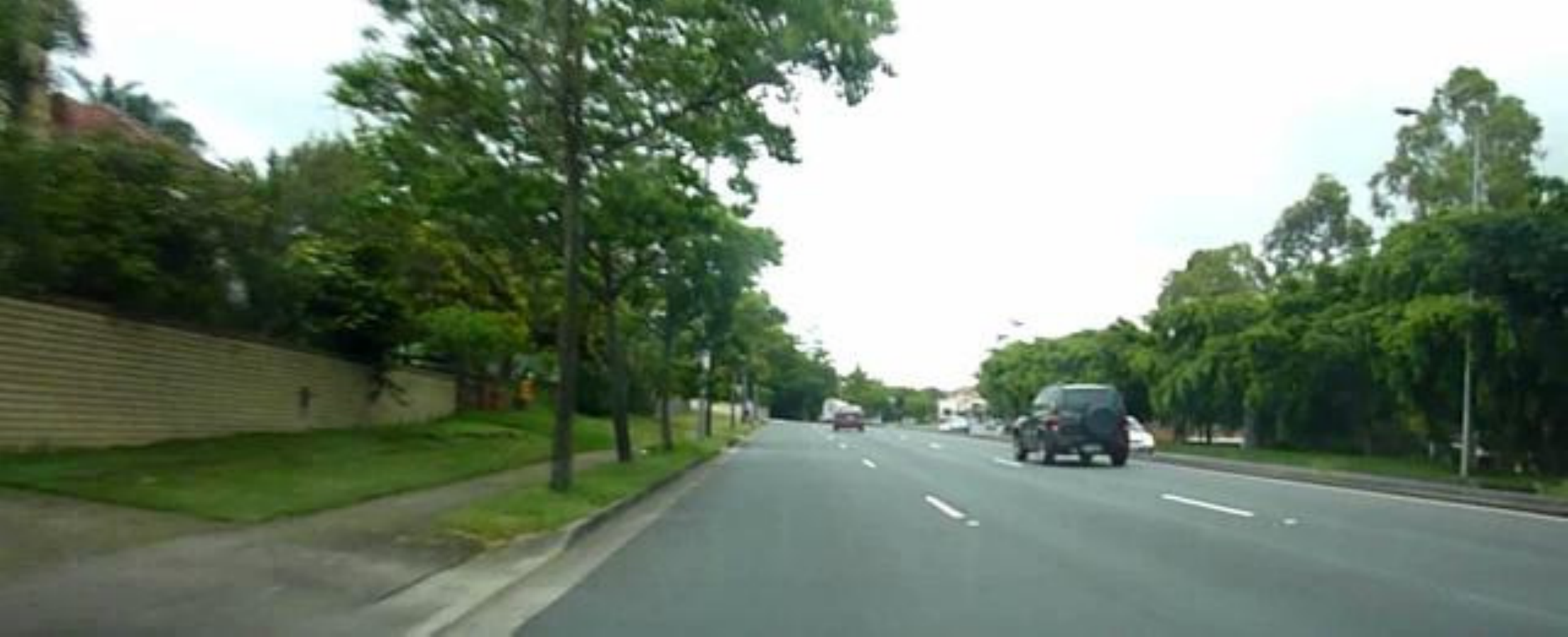}
	}
	\caption{An example image pair from the Alderley dataset. Note that these frames are extremely difficult to match, even for a human.}
	\label{fig:alderley_ex}
\end{figure}

\begin{figure}[!t]
	\centering
	\includegraphics[width=\columnwidth]{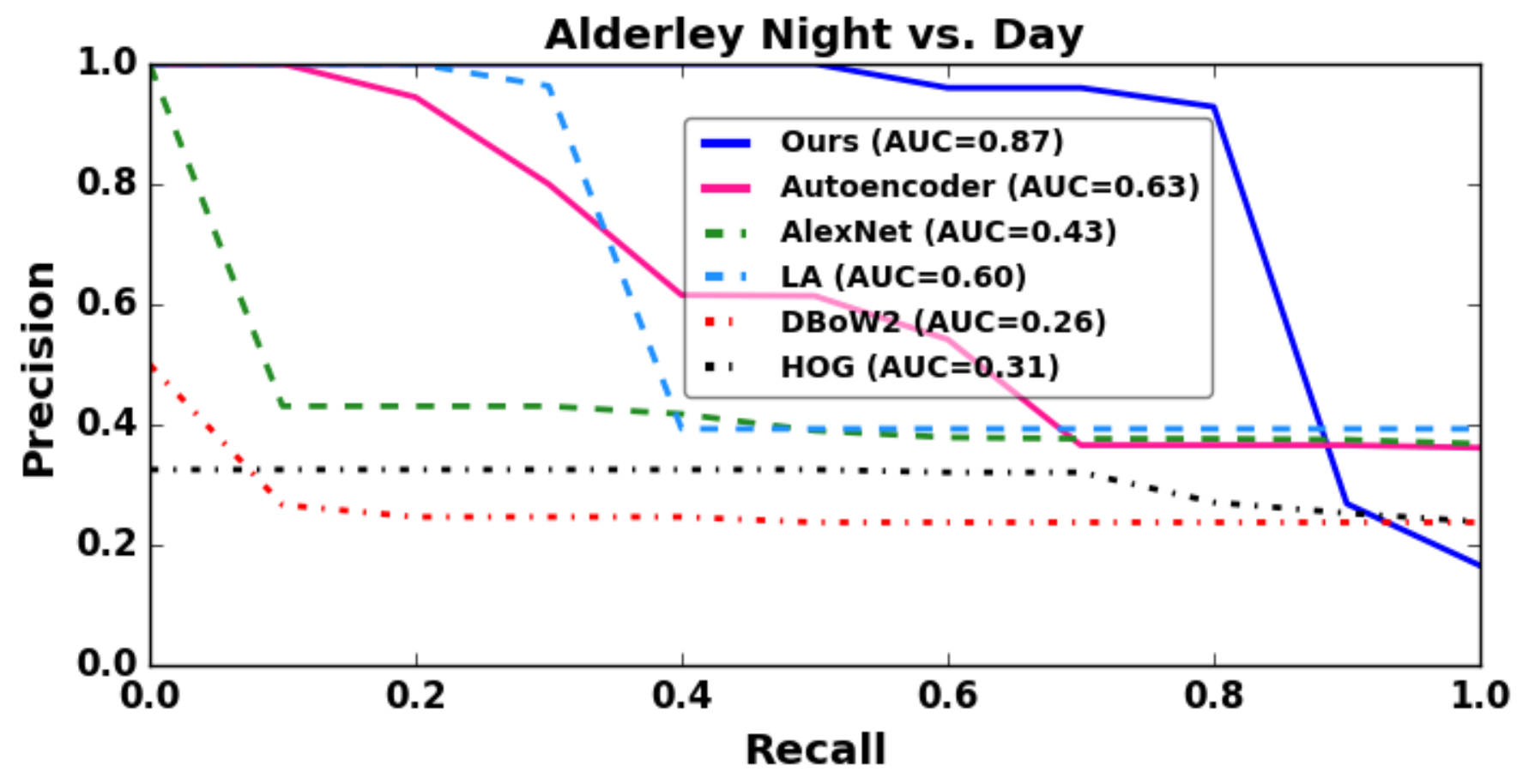}
	\caption{Our method outperforms the state-of-the-art algorithms on the {Alderley dataset}, with the highest AUC and $r$ value.}
	\label{fig:alderley_night_v_day}
\end{figure}

\subsection{The Alderley Dataset}

The {Alderley dataset} %\footnote{\url{https://wiki.qut.edu.au/pages/viewpage.action?pageId=181178395}} 
was first introduced in SeqSLAM~\cite{SeqSLAM} 
and is composed of two image sequences, extracted from videos taken during a rainy night and a sunny day. 
Fig. \ref{fig:alderley_ex} shows an example image match from this dataset; it is very difficult even for human to realize these images are of the same place.
Frame correspondences are included in the dataset, 
providing ground truth for place recognition, with an added tolerance for multiple sequential frames of the same location.
We test on the last 200 frames of each sequence.
The comparison results are shown in Fig. \ref{fig:alderley_night_v_day}.
Clearly, our method is the most robust in this case, taking the highest AUC and $r$ value by large margins.
Interestingly, the regular autoencoder performs well here.
Note that the model from \cite{LOPEZANTEQUERA2017} was trained on a different subset of the Alderley dataset than used here, giving their model an advantage over the others that have not been trained on any of the Alderley dataset. 
Nevertheless, our model is still more robust in this experiment, 
while the other methods are failing due to the significant differences in appearance in this dataset.

\subsection{The Gardens Point Dataset}

The {Gardens Point dataset} 
consists of three traversals through the QUT campus in Brisbane, Australia. 
In this dataset, there are two day-time traversals -- 
one  tends to contain images of the left side of the walking path, and the other contains the right side. 
Additionally, there is one night-time traversal, which tends to the right side of the path as well.
Unlike in the Alderley dataset, image $i$ from one sequence matches image $i$ from any of the other two.
We utilize this, as well as an added tolerance for multiple sequential images of the same location, to define the ground truth for this experiment in addition to the remaining precision-recall experiments in this work, since the rest of the datasets follow the same format.

\begin{figure}[!t]
	\centering
	\includegraphics[width=\columnwidth]{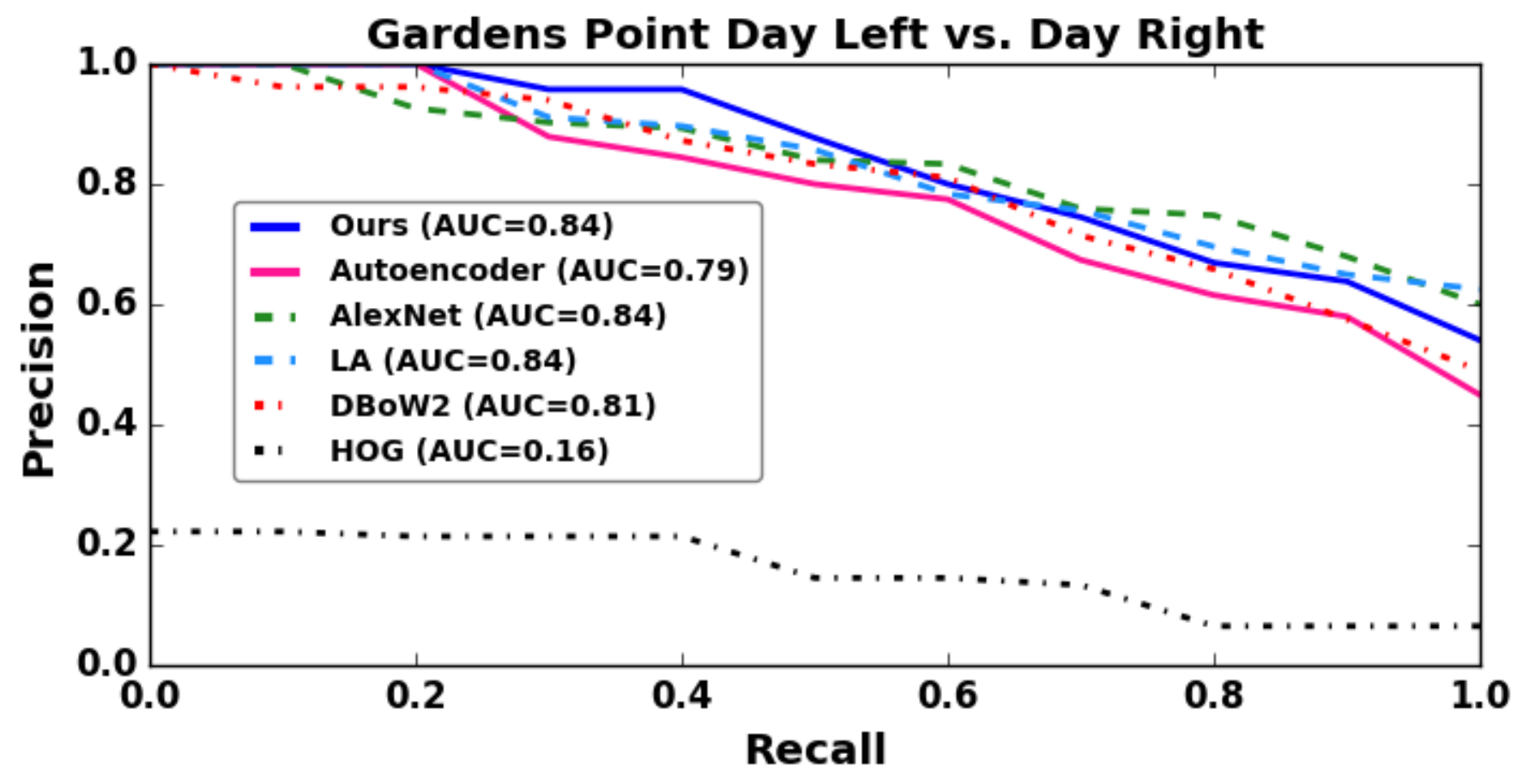}
	\includegraphics[width=\columnwidth]{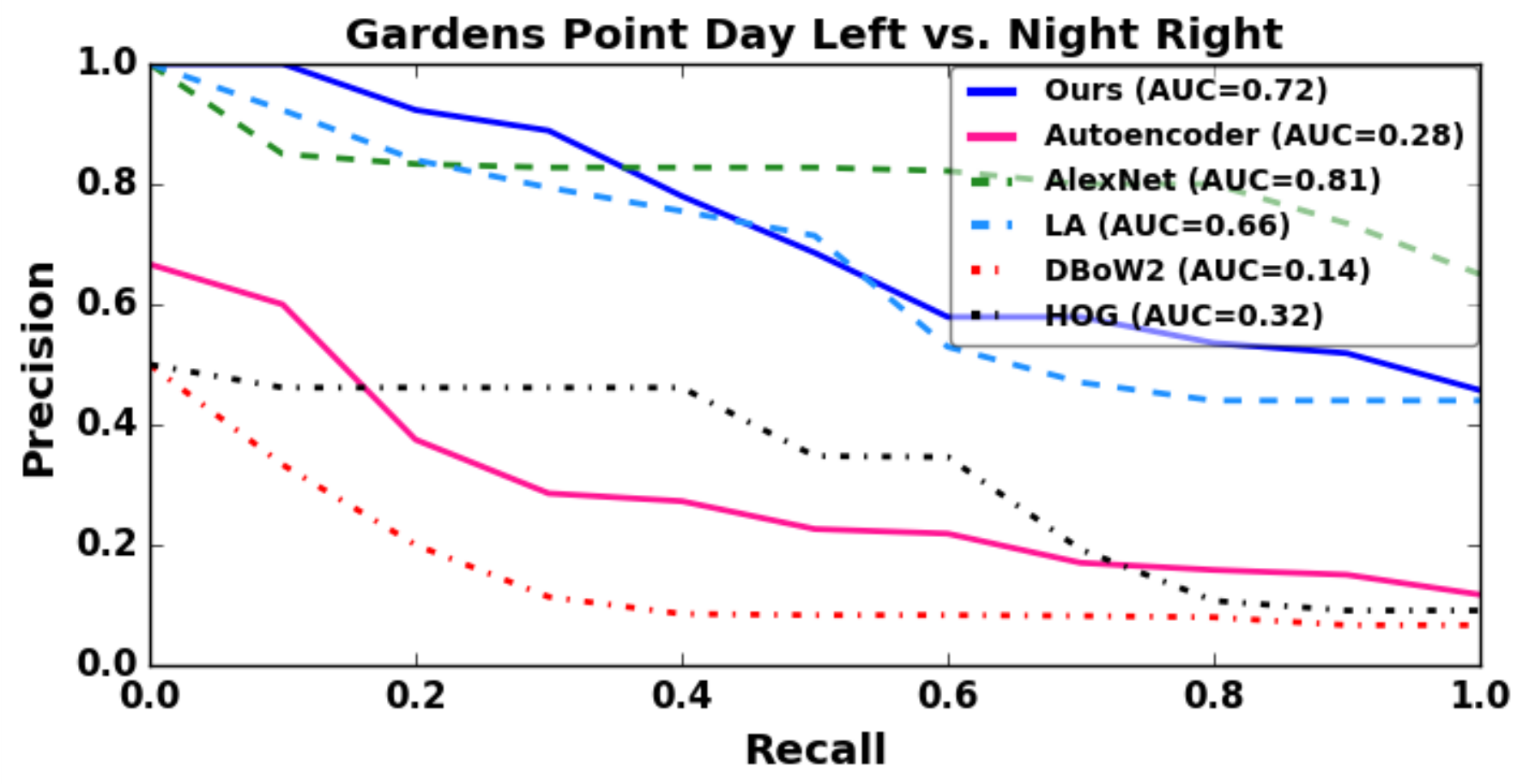}
	\caption{The comparison results on the Gardens Point dataset. 
	(top) Our method performs comparably with \cite{LOPEZANTEQUERA2017} (which, however, is a supervised learning approach) in the {\em  day-time sequence}, while (bottom) our method  outperforms its competitors  in the {\em night-time sequence}.
	}
	\label{fig:gp_day_left_v_right}
\end{figure}

An example of this dataset is shown in Fig.~\ref{fig:gp_example},
while Fig. \ref{fig:gp_day_left_v_right} shows the comparative results. 
Fig. \ref{fig:gp_day_left_v_right} (top) is for the {\em day left and day right sequences}.
Our method, AlexNet, and LA perform comparably in this dataset, and even the autoencoder is not far behind; however, we will see that this trend of comparable performance does not carry throughout the experiments.
DBoW2 is competitive in this experiment, but falls short of our method, AlexNet, and LA.
The HOG descriptor we used to train our model is clearly not nearly as robust as our final descriptor, even in this daytime dataset -- one of the easier datasets used in experimentation.
To further challenge these methods, 
we use the {\em night-time sequence} from the Gardens Point dataset -- 
the results of which are shown in Fig. \ref{fig:gp_day_left_v_right} (bottom). 
In this case, our method takes the highest $r$ value, and the second-highest AUC. 
DBoW2, HOG, and the autoencoder completely fail in this test.
Although the performance of DBoW2 could most likely could be improved by training the ORB vocabulary tree on the night-time images,  
we want to test all of these methods void of any environment-specific training for the purpose of generalization.

\subsection{The Nordland Dataset}

\begin{figure}[!t]
	\centering
	\subfloat{
	  \centering
	  \includegraphics[width=.46\columnwidth]{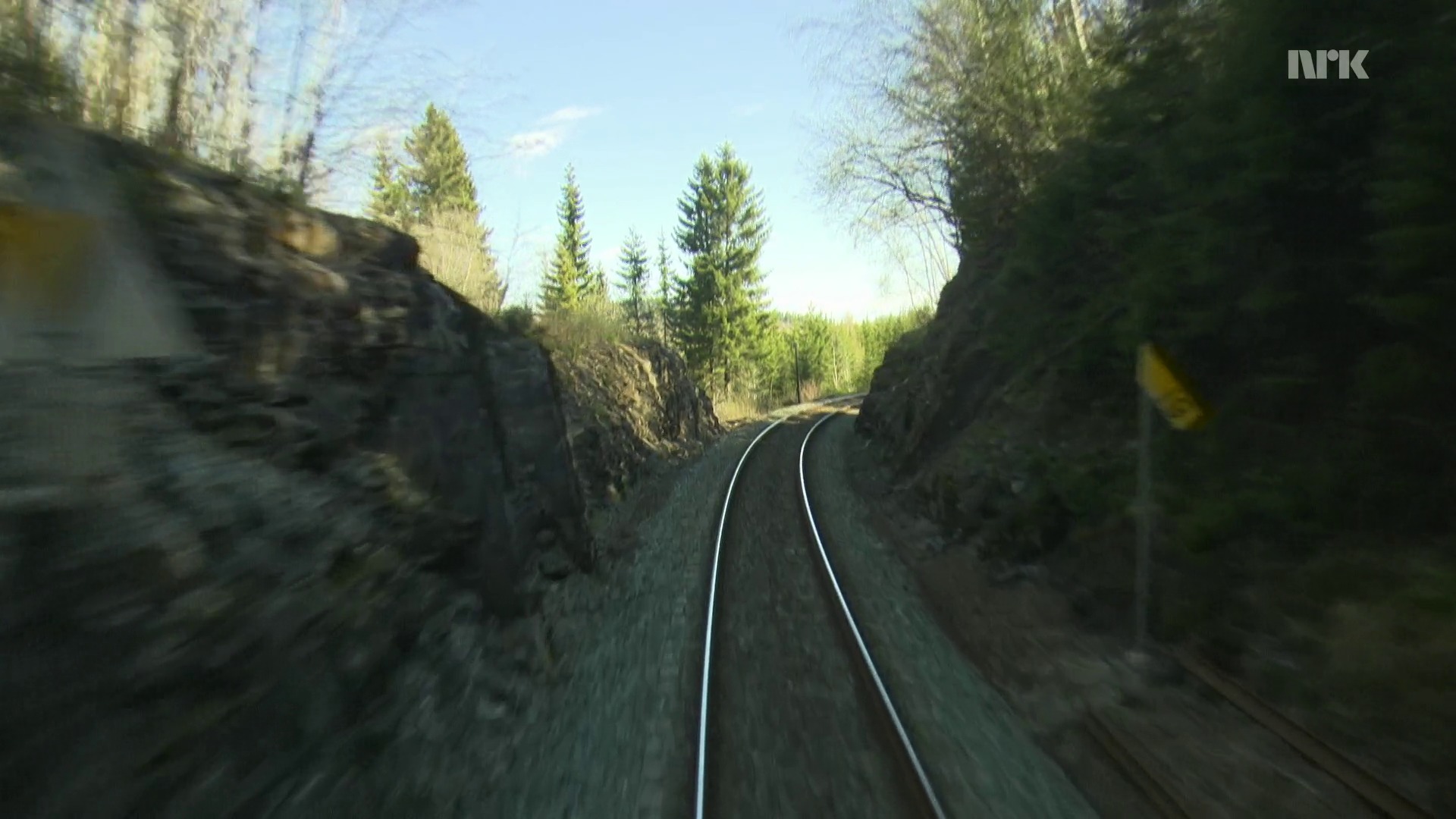}
	} %\quad
	\subfloat{
	  \centering
	  \includegraphics[width=.46\columnwidth]{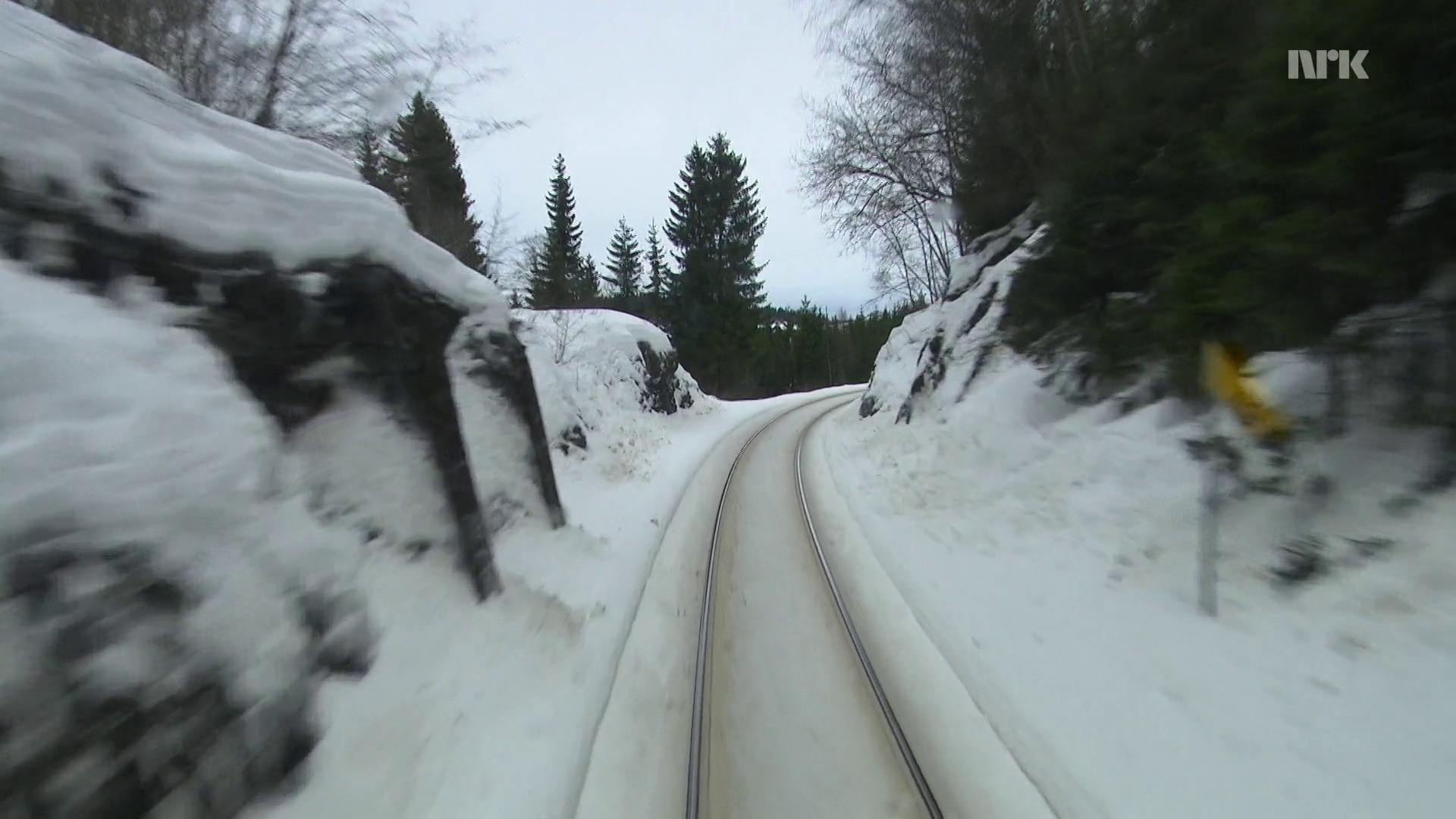}
	}
	\caption{An example image pair from the Nordland dataset. The left image is from the spring sequence while the right one is from the winter.} 
	\label{fig:nord_ex}
\end{figure}

The {Nordland dataset},  
one of the most challenging place recognition datasets to date~\cite{KENSHIMOV2017124},
consists of four time-synchronized videos of train journeys through Norway.
Each of the four 9-hour long sequences corresponds to a different season, creating a difficult challenge for cross-seasonal place recognition.
In addition to seasonal variation, the images also contain extreme blurring from the fast speed of the train.
Fig. \ref{fig:nord_ex} shows an example of an image pair from this dataset.
We test our method on one of the most difficult sequence pairs, {\em Winter versus Spring}.
Specifically, we extracted 5,357 frames from these two videos.
This experiment was performed using frames 29 to 200, as this was the first sequence we found where the train was constantly in motion and outside of tunnels.
 Note that images from inside the tunnels are completely black, and therefore useless for experimentation; additionally, if the train was stopped at a station, there were too many sequential images of the same location, causing large biases in the precision-recall curves.

\begin{figure} [!t]
	\centering
	\includegraphics[width=\columnwidth]{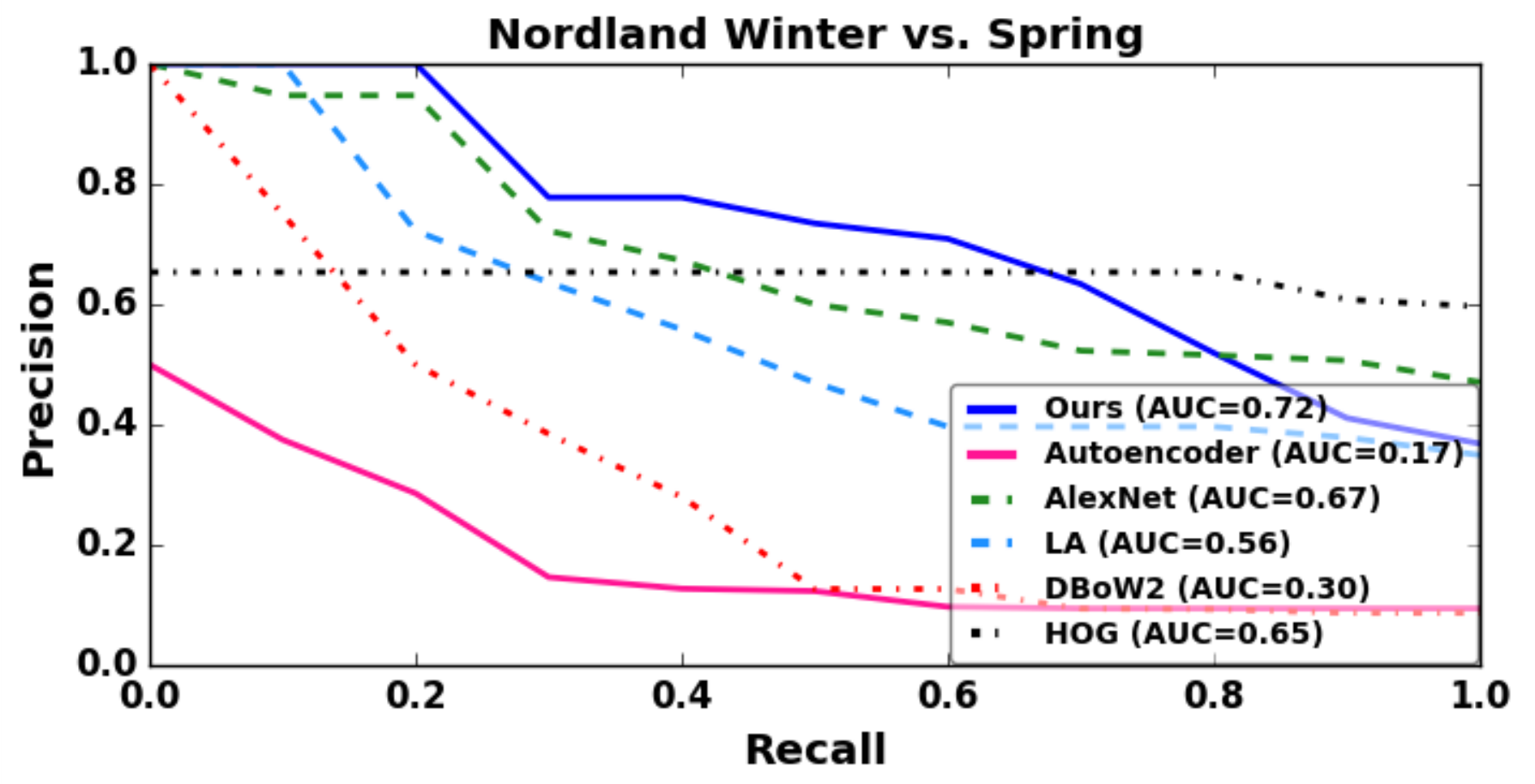}
	\caption{Comparison results on the Nordland dataset.
	Our method is observed to be more robust to the seasonal changes provided by this subset of the winter and spring sequences.
	}
	\label{fig:nord_spring_v_winter}
\end{figure}

The results of this experiment are shown in Fig.~\ref{fig:nord_spring_v_winter}.
It should be noted that \citet{LOPEZANTEQUERA2017} used all but the last hour of each Nordland sequence in training their model; this implies that their model has seen this testing data in the training phase.
However, even with this incredible disadvantage, our model outperforms theirs, along with other methods in this experiment.


\subsection{Our Campus Loop Dataset}

\begin{figure}[!t]
	\centering
	\subfloat{
	  \centering
	  \includegraphics[width=.475\columnwidth]{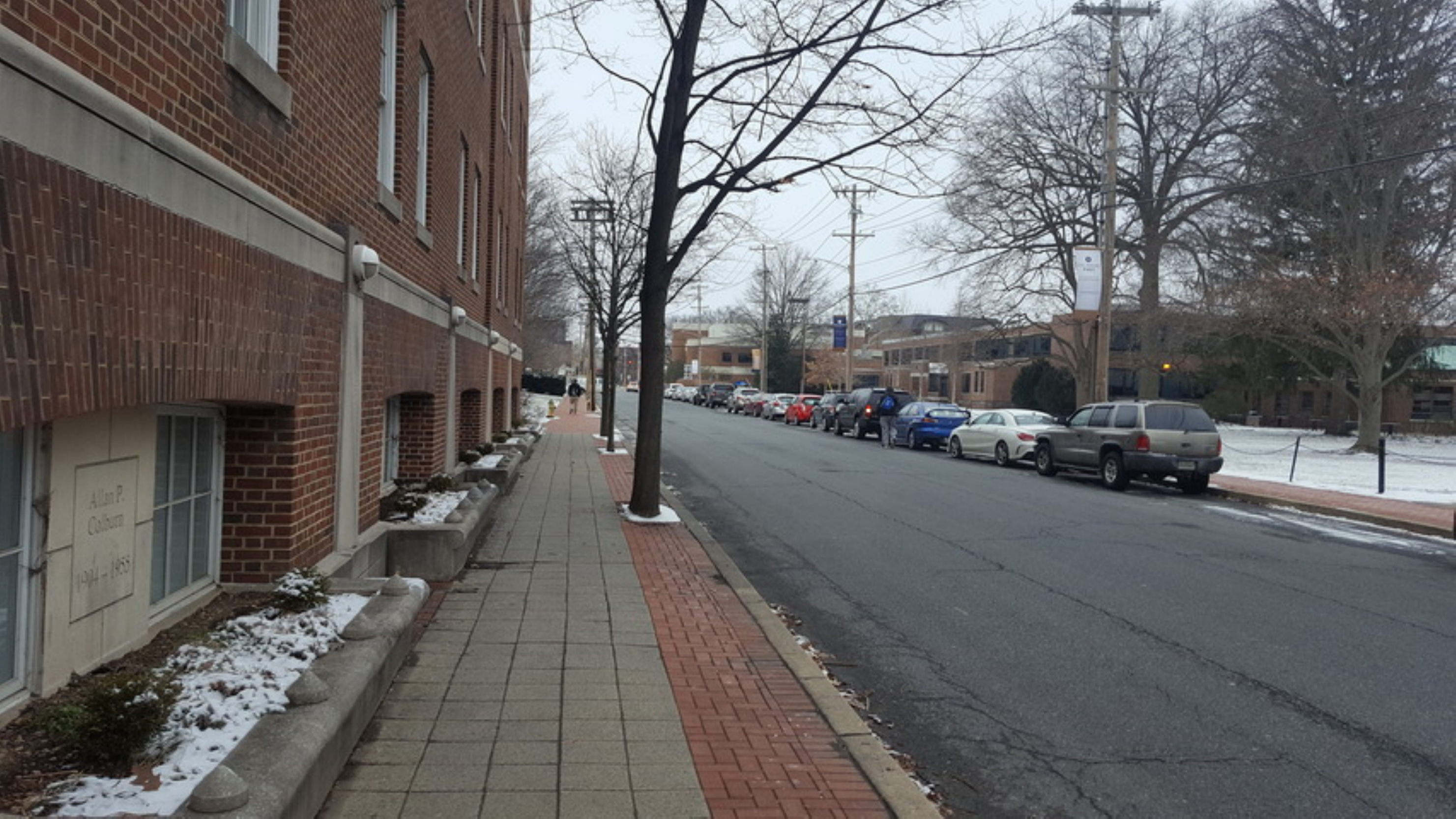}
	  \label{fig:campus_loop_ex:subfig1}
	} %\quad
	\subfloat{
	  \centering
	  \includegraphics[width=.475\columnwidth]{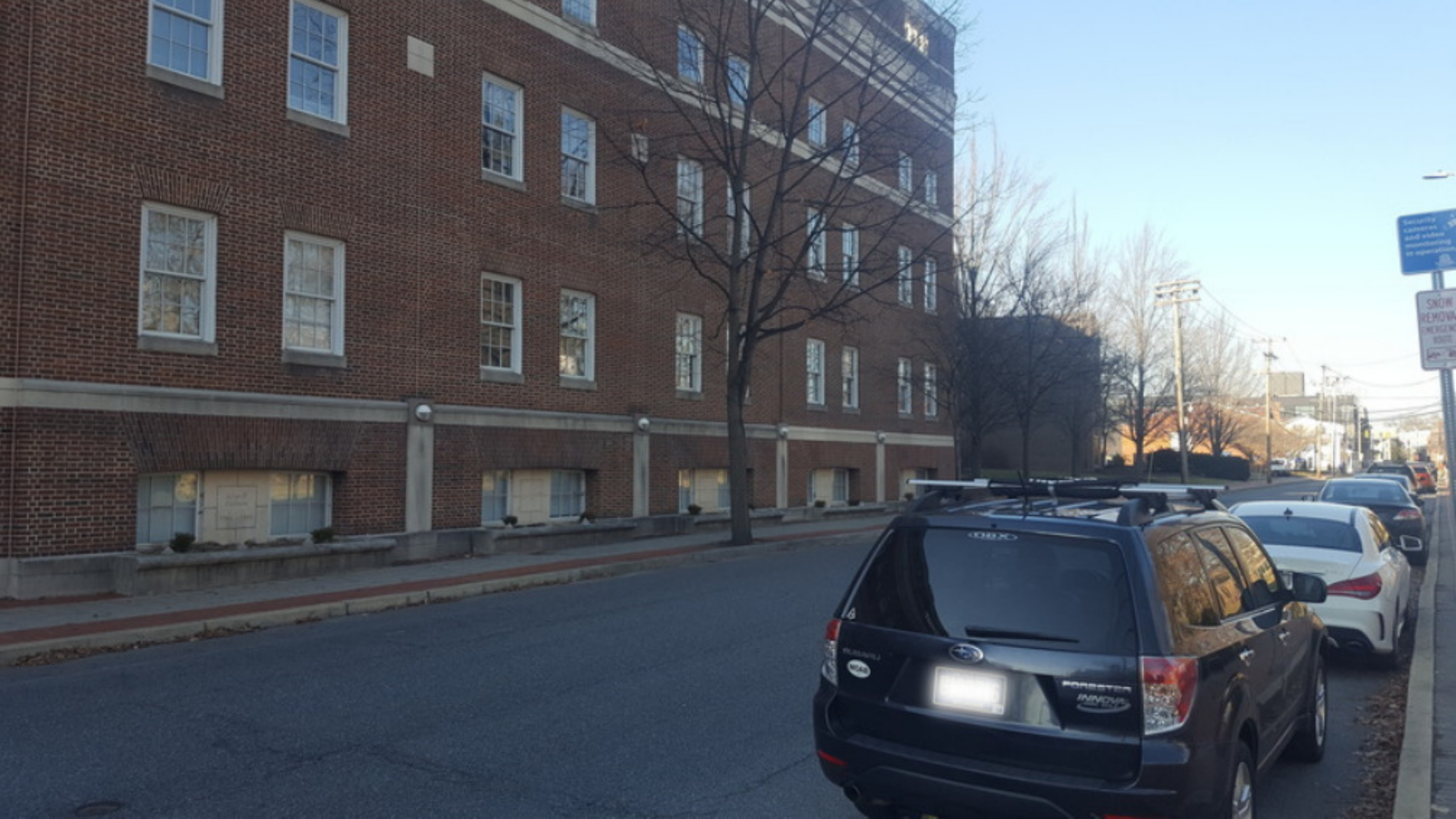}
	  \label{fig:campus_loop_ex:subfig2}
	}
 	\caption{An image pair example from our Campus Loop dataset, which has extreme variations in viewpoint, weather, and dynamic objects.}
	\label{fig:campus_loop_ex}
\end{figure}

The Nordland dataset provides extreme weather variations, the Gardens Point dataset provides extreme brightness and viewpoint variations, as well as many dynamic objects,  while the Alderley dataset provides all but large viewpoint variations.
However, we found that no dataset can provide all of these challenges.
Therefore, we collect our own dataset, termed as the {\em Campus Loop dataset}. 
The dataset consists of two sequences of 100 images each.
The sequences are a mix of indoor and outdoor images in a campus environment.
The first sequence was taken on a snowy day, when it was very cloudy, while the second was taken nine days later, when most of the snow had melted and the sun was out.
The indoor images obviously do not vary as much with this weather change.
Additionally, each image match contains varying perspectives and many dynamic objects, making this one of the most challenging publicly-available place recognition datasets.
Fig. \ref{fig:campus_loop_ex} shows an example of an image pair from this dataset.

\begin{figure}[!t]
	\centering
	\includegraphics[width=\columnwidth]{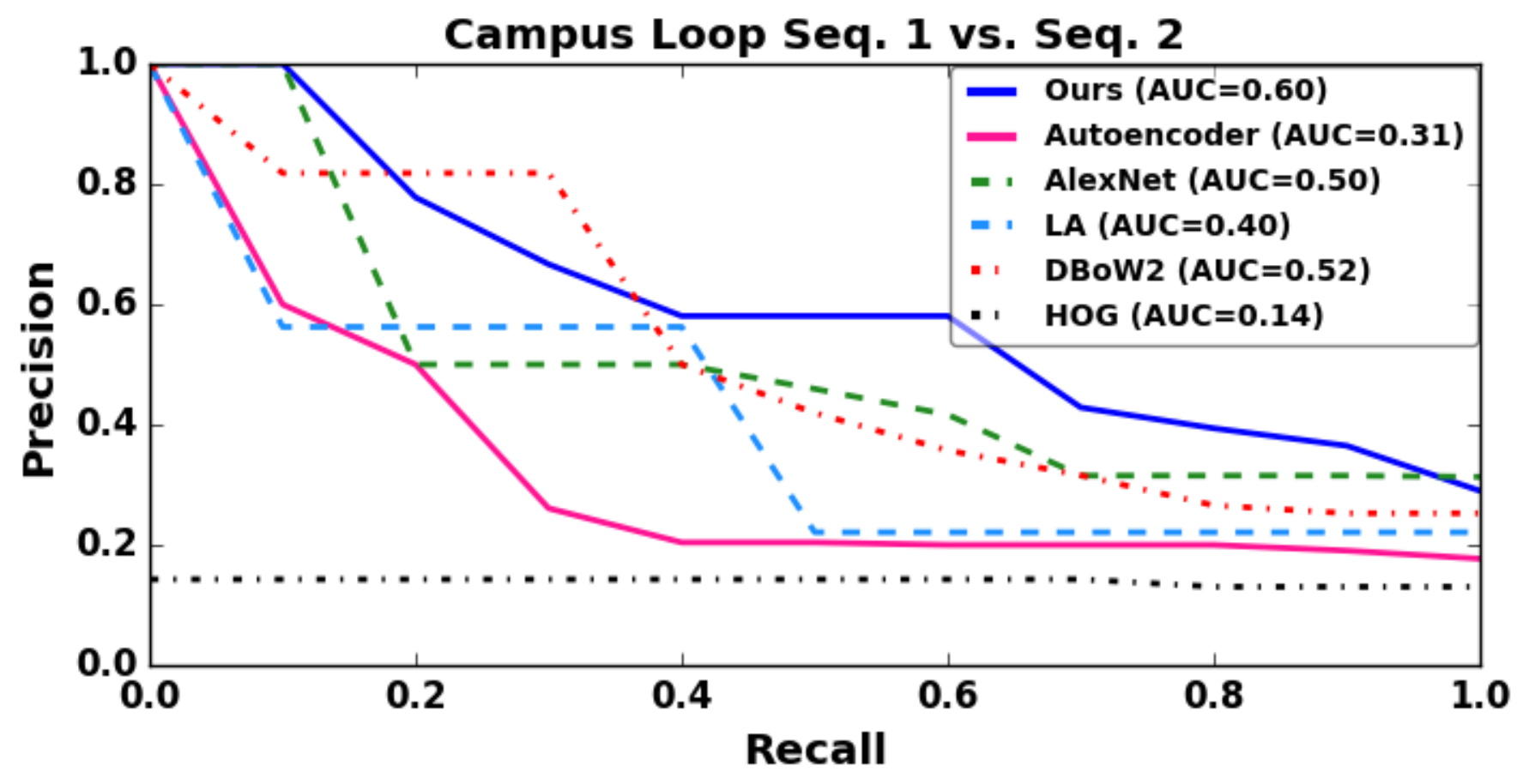}
	\caption{Our approach outperforms the other benchmark methods on our own Campus Loop dataset,
	with the highest $r$ value while tying with AlexNet \texttt{conv3} for the highest AUC.}
	\label{fig:campus_loop}
\end{figure}

The results of experimentation with this dataset are shown in Fig. \ref{fig:campus_loop}.
As expected, across the board the performance is worse than any other dataset thus far.
Nevertheless, comparatively, 
our method is the most robust to the challenges presented in this new dataset.
The model from  \citet{LOPEZANTEQUERA2017} falls flat in this experiment, 
performing significantly worse than the other three deep-learning methods, and falling short of even DBoW2 in AUC.

%%%
\subsection{Runtime Evaluation}
\label{sec:runtime-eval}

To validate the efficiency of our approach,
we perform runtime evaluations of both descriptor computing time, and database querying time for a single-nearest neighbor search.
These tests are conducted on affordable hardware to allow for better reproducibility -- 
specifically, an i7-6700HQ CPU, and a GeForce GTX 960M GPU.
Note that in this test we only compare against DBoW2 and AlexNet due to the following:
(i) \citet{LOPEZANTEQUERA2017} do not provide an open-sourced library for performing image matches, 
so it would not necessarily be fair to time their models in our code.  
They report 1.8 millisecond descriptor computing time using a GPU, which is slower than ours.
However, their descriptor is smaller than ours, so it should be cheaper to query, while ours is shown to be competative (if not better) in accuracy, and more convenient to fine-tune or retrain.
%
%(ii) The 64,896-dimensional AlexNet \texttt{conv3} is too large of a descriptor to be competitive with ours 
%in terms of speed without approximating the cosine similarity score \cite{Ravichandran2005} or applying a dimensionality reduction as done for the precision-recall curves in this work.
%
(ii) The reduced HOG is presented in the preceding tests only to show that our model learns a better version of it, rendering its runtime irrelevant.

We choose DBoW2 with ORB features as one benchmark, since it is one of the fastest place recognition libraries used in many state-of-the-art SLAM systems (e.g., \cite{murORB2, murTRO2015}).
Additionally, we choose AlexNet \texttt{conv3} features both with GRP, compressed to 1,064 dimensions, and in original form, since it is a popular choice for ConvNet-based place recognition (i.e. in \cite{sunderhauf_place_2015, Hou2017, BAI2018, KENSHIMOV2017124}).
Note that AlexNet has been modified to only contain up to the \texttt{conv3} layer here for fair testing.
The {\em KITTI Visual Odometry} dataset sequence 00 \cite{Geiger2012CVPR} is used as testing data for the first experiment.
We utilize the 4,541 $376 \times 1241$ stereo pairs from this sequence to construct two subsets, placing all of the left images in the database, and using the right images for querying.
Table \ref{table:exec_time} shows the results of this experiment,
where feature extraction time refers to the time between starting with a raw image and ending with having that image's representation inserted into the respective database.
%So for DBoW2, this includes the ORB extraction time,
%which is a fair comparison to our method that requires no intermediate image representation.
%
The query times do not include any descriptor calculation times.
%Note that we have {\em no} query times for GPU, since we always copy our descriptors back to RAM in order to conserve precious GPU memory.
Note that DBoW2 has no GPU implementation, and AlexNet with GRP produces features of the same size as ours, so the query times will be the same.
From Table \ref{table:exec_time}, it is clear that our method is faster than the others for feature extraction when a GPU is used to make forward passes through the net, and is still reasonably fast when using a CPU.
Additionally, our method for querying, though it is simple, outperforms DBoW2 in terms of speed in this experiment.

\begin{table}[!t]
	\caption{Times (in milliseconds) to extract features and query a database of 4,541 images on the KITTI dataset.} 
	\centering
	\resizebox{0.5\textwidth}{!}{
	\begin{tabular}{@{\extracolsep{4pt}}lllllll}
		\toprule   
		 Method  & \multicolumn{2}{l}{Extract (GPU)} & \multicolumn{2}{l}{Extract (CPU)} & \multicolumn{2}{l}{Query} \\ 
		         & $\mu$ & $\sigma$                          & $\mu$ & $\sigma$                          & $\mu$ & $\sigma$ \\
		\midrule
		 Ours    & \bf{0.862} & 0.025                    & 44.0  & 2.98                              & \bf{1.47}  & 0.031 \\
		 DBoW2   & N/A        & N/A                          & 15.8  & 3.08                              & 4.25  & 0.547 \\
		 AlexNet (no GRP)& 2.13  & 0.038                             & 405.0   & 17.4                              & 80.8  & 0.708 \\
        AlexNet & 16.6  & 0.658                             & 418.0   & 17.8                              & N/A & N/A \\
		\bottomrule
	\end{tabular}
	\label{table:exec_time}
	}
\end{table}

\begin{figure}[!t]
	\centering
	\includegraphics[width=\columnwidth]{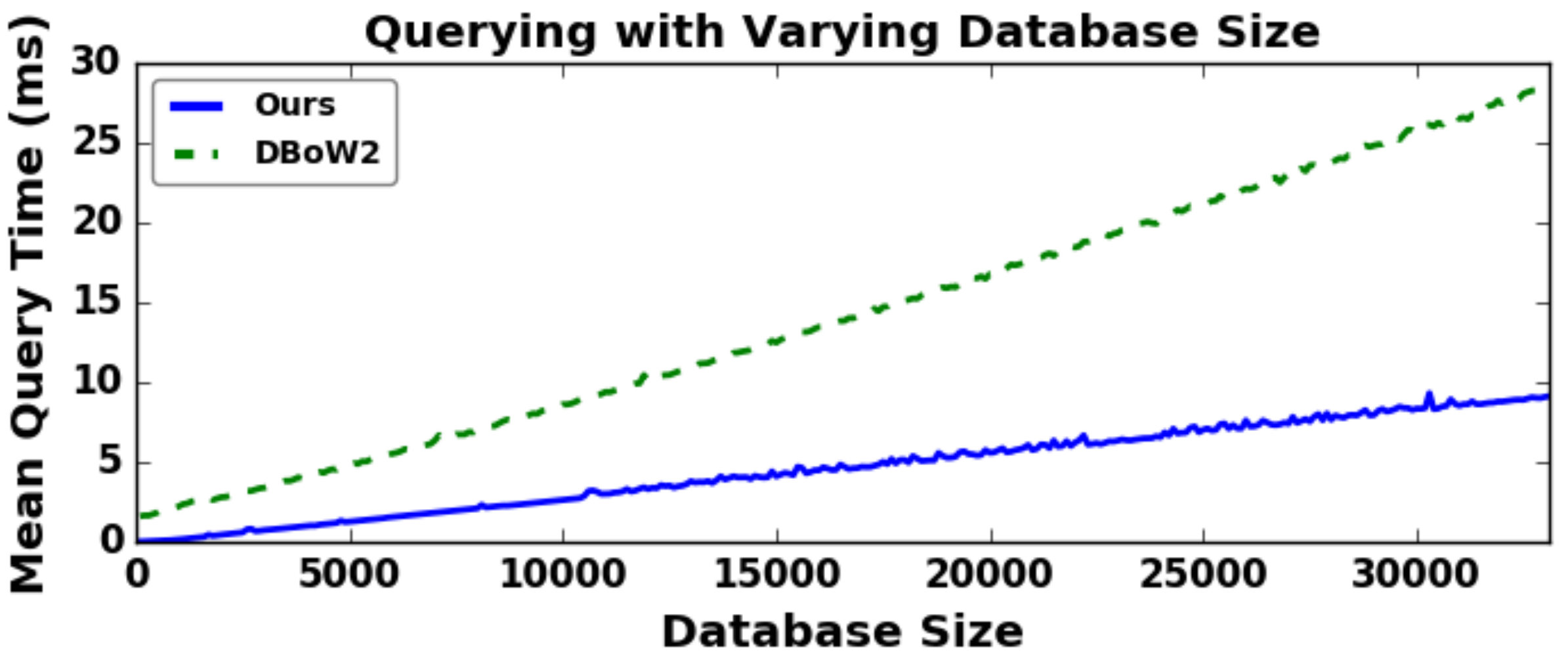}
	\caption{The proposed method performs queries faster than DBoW2 with varying database size.}
	\label{fig:query_time}
\end{figure}

We also test the query speed for a variable-sized database, comparing only to DBoW2 -- the most competative candidate from Table \ref{table:exec_time}.
We use the large {\em St. Lucia} dataset~\cite{Warren2010}, which, similar to KITTI, is a sequence of stereo pairs. 
However, this sequence contains over 30,000 stereo pairs, making it very useful for testing a variable database size.
The left images are used for the database, and a subset of the right images is used for querying. % -- using every tenth of the first 1,000.
%We use the left images to incrementally construct a database, timing the queries every time we add 100 of them.
%
Fig.~\ref{fig:query_time} shows the results of this experiment,
from which it is evident that our querying method is inexpensive, even for very large databases
 -- larger than that created by a typical SLAM system.

\subsection{Online Loop Closure} \label{sec:online}

\begin{figure}[!t]

	\centering
	\subfloat{
	  \centering
	  \includegraphics[width=0.75\columnwidth]{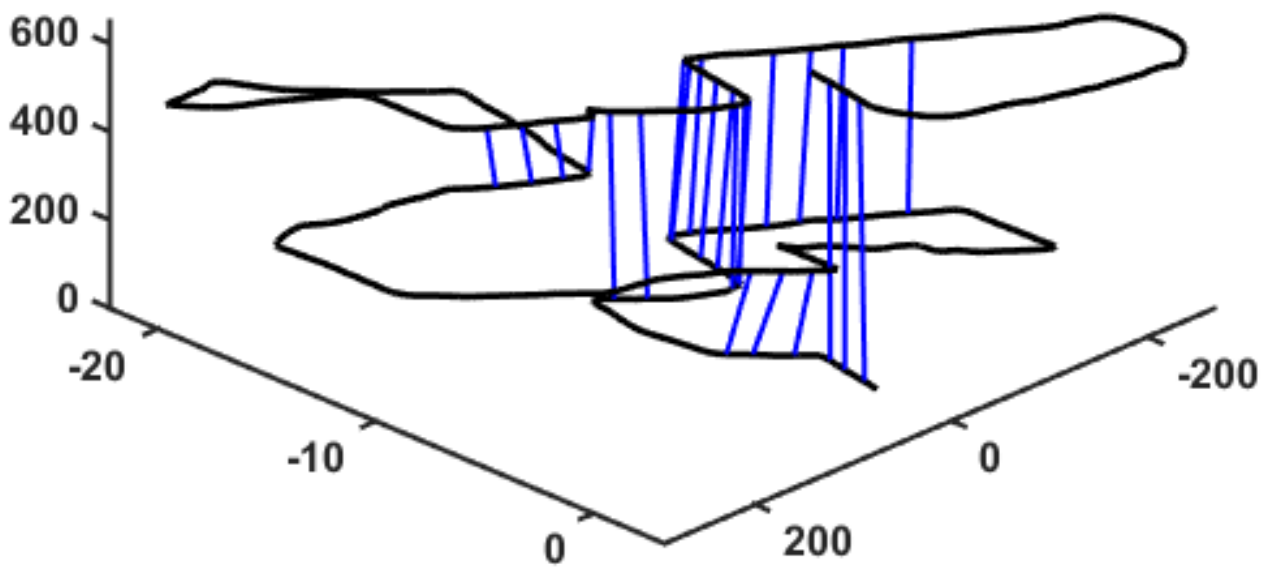}
	  \label{fig:kitti_loops:00}
	} \quad
	\subfloat{
	  \centering
	  \includegraphics[width=0.75\columnwidth]{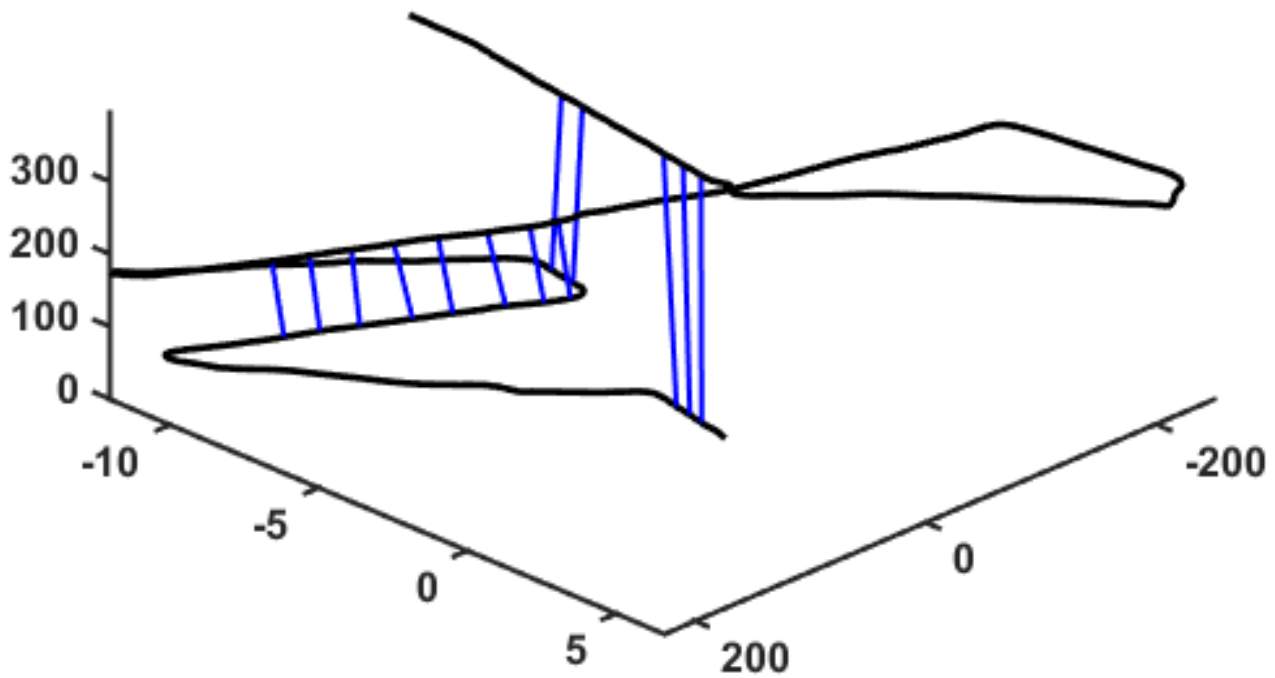}
	  \label{fig:kitti_loops:05}
	}
	\caption{
		The results of online loop closure using KITTI 00 and 05, respectively.
		The 2D location of the trajectory is represented on the $x$-$y$ plane, and the $z$-axis is the current keyframe number.
	}
	\label{fig:kitti_loops}

\end{figure}

Precision-recall curves are a good metric for binary classification, but they do not fully prove that our method is capable of accurately closing loops in practice.
Therefore, we perform real-time loop closure using an extremely simple application of our model on KITTI~\cite{Geiger2012CVPR} sequences 00 and 05.
In this experiment, we simulate keyframe selection by using every seventh frame for loop detection.
A loop closure hypothesis is proposed if a database query score is above an a-priori threshold $\tau$, and a loop is determined closed if three consecutive queries retrieve descriptors within six frames of the first query.
We exclude the most recent images from the search space, and do not start loop detection until the database is of sufficient size.
%During every query we exclude the most recent 77 frames from the search space, and do not start querying until the database is of size 77.
We choose $\tau$ from the precisions and recalls shown in Fig.~\ref{fig:gp_day_left_v_right} (bottom) such that it maximizes the recall rate with perfect precision.
Fig.~\ref{fig:kitti_loops} shows the results of this experiment.
Clearly our method is able to consistently close loops on a practical SLAM dataset using a threshold from completely unrelated ground truth data, which shows that it is ready to use in a real-time SLAM system.
Additionally, this application of our model for online loop closure is extremely simple, and can be improved upon easily by looking at the $k$-nearest neighbors instead of the single-nearest neighbor, adding extra false positive rejection methods (i.e. a geometric check), or utilizing any of the methods described in the next section.

\subsection{Integration into ConvNet-Based Place Recognition Systems} \label{sec:improving_performance}

As stated in Section \ref{sec:online_use}, our model can easily be integrated into off-the-shelf ConvNet-based place recognition systems~\cite{sunderhauf_place_2015, KENSHIMOV2017124, Hou2017, BAI2018} for faster feature extraction.
These methods build upon the use of holistic image descriptors, improving performance in different cases.
They treat the ConvNet as a black box for image description -- throwing out image classifications from classification networks.
Many of these methods are forced to reduce the dimension of the ConvNet features to minimize runtime, while
our model already produces a small enough descriptor for real-time use, and is smaller and faster than the typical classification network used by these methods.

\begin{figure}[!t]
	\includegraphics[width=\columnwidth]{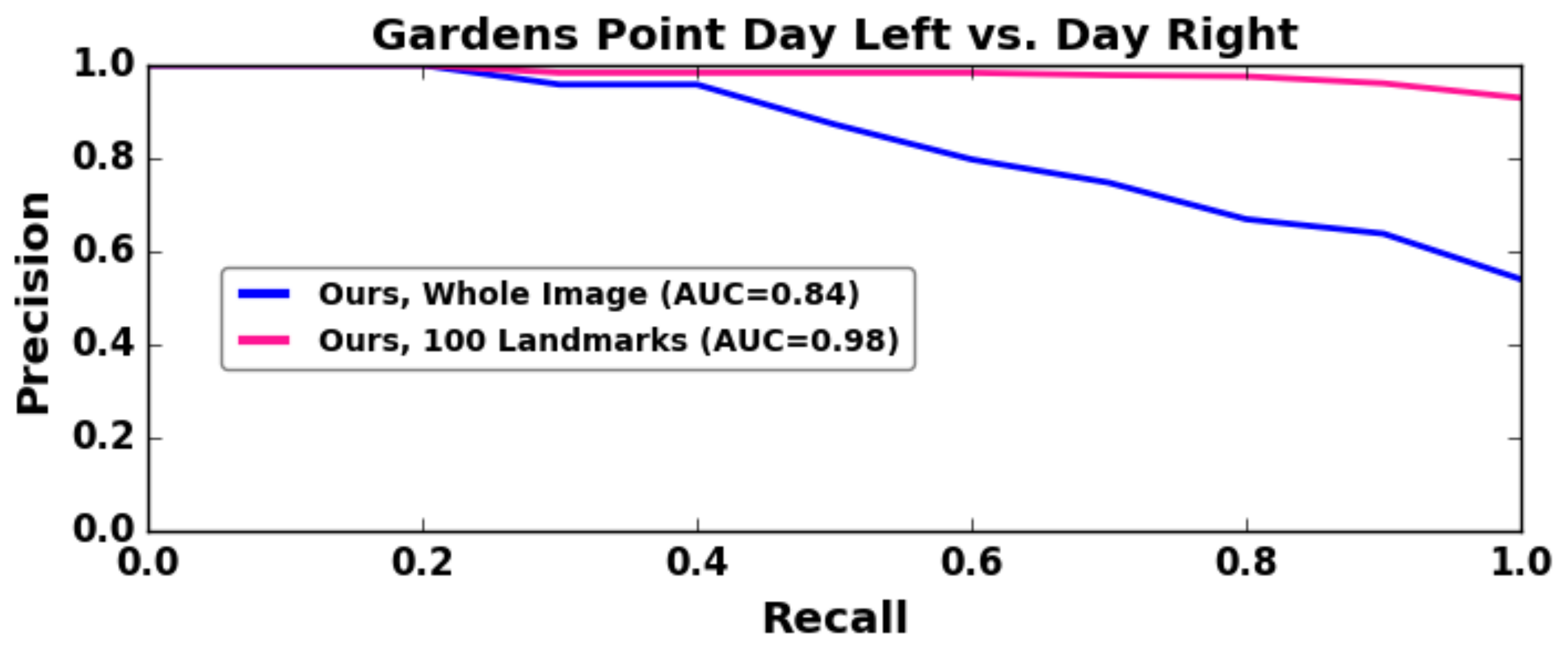}
	\caption{The landmark-based method shows a vast improvement over the holistic approach -- nearing perfect performance.}
	\label{fig:landmark_method}

\end{figure}

To prove this, we reproduce the state-of-the-art landmark-based place recognition system presented in~\cite{sunderhauf_place_2015}, replacing Edge Boxes~\cite{Zitnick2014EdgeBoxes} with BING~\cite{Cheng2014BING} as \citet{Hou2017} did to reduce runtime, and replacing the dimension-reduced AlexNet \texttt{conv3} landmark descriptor with that from our model.
\citet{sunderhauf_place_2015} proposed reducing the AlexNet \texttt{conv3} layer to 1,024 dimensions, while ours is naturally 1,064, so we do not need to reduce it further -- avoiding the cost of the $ 1024 \times 64896 $ by $ 64896 \times m $ matrix multiplication required to project $m$ landmark descriptors (using AlexNet) into 1,024 dimensions, which must be done every time a new image is added to the database.
The results of this experiment can be seen in Fig.~\ref{fig:landmark_method}.
The landmark-based method offers an enormous improvement over the holistic image descriptor -- approaching perfect performance on the Gardens Point daytime dataset.
Our model is seamlessly integrated into this system, which suggests that it can easily replace bloated classification networks in other such ConvNet-based place recognition systems.

\section{Conclusions} \label{sec:conclusions}

We have presented a novel unsupervised deep neural network for fast and robust loop closure, applicable in visual SLAM.
Built upon the denoising autoencoder architecture, we apply randomized projective transformations to images 
in order to capture extreme variations in viewpoints due to robot motion,  
while employing the fixed-length HOG descriptor to help our network better learn the geometry of scenes. 
The proposed model allows for vast amounts of data to be used in training, since none of it needs to be labeled or contain any special information.
Furthermore, although our pre-trained model generalizes well in its current state, it is easy to fine-tune or retrain due to our unsupervised design -- increasing the likelihood of improvement as more data becomes available.

We have performed thorough comparison studies on different datasets against the state-of-the-art image description methods for place recognition,
where the extensive experimental results have shown that the proposed deep loop closure method generally outperforms the benchmarks in terms of both effectiveness (precision-recall) and efficiency (runtime).
Our model is compact, robust, and fast -- making it a promising candidate to replace larger, slower classification networks in ConvNet-based place recognition systems, as we have shown by reproducing \cite{sunderhauf_place_2015}.
Due to its lightweight yet robust design, our model is suitable for use in real-time SLAM systems -- in particular, direct algorithms~\cite{dso, lsd_slam, dtam, direct_vins} where no intermediate image representation is needed.
We aim to provide an out-of-the-box solution for loop closure, and, more generally, place recognition.
We are currently working to integrate our model into various SLAM systems, applicable for autonomous navigation in challenging environments.

%% Bib %%%%%%%%%%%%%%%%%%%%%%%%%%%%%%%%%%%%%%%%%%%
\bibliographystyle{IEEEtranN}  
\bibliography{library}

%%%%%%%%%%%%%%%%%%%%%%%%%%%%%%%%%%%%%%%%%%%%%
\end{document}